\pdfoutput=1

\documentclass[11pt]{article}

\usepackage[final]{acl}

\usepackage{times}
\usepackage{latexsym}

\usepackage[T1]{fontenc}

\usepackage[utf8]{inputenc}

\usepackage{microtype}

\usepackage{inconsolata}

\usepackage{graphicx} 
\usepackage{amsmath}
\usepackage{amssymb}
\usepackage{booktabs}
\usepackage{multirow}
\usepackage{makecell}
\usepackage{caption}
\usepackage{subcaption}
\usepackage{xspace}
\usepackage{pifont}
\usepackage[most]{tcolorbox}

\newcommand{\data}{\textbf{\texttt{MAQA}}\xspace}

\newcommand{\cmark}{\ding{51}} %
\newcommand{\xmark}{\ding{55}} %

\def\ie{\emph{i.e}.\xspace}

\newtheorem{observation}{Observation}

\title{MAQA: Evaluating Uncertainty Quantification in LLMs \\  Regarding Data Uncertainty}

\author{Yongjin Yang${}^{1}$\thanks{This work was done during the internships at NAVER AI Lab.} \quad
  Haneul Yoo${}^{1}$\footnotemark[1] \quad
  Hwaran Lee${}^{2,3}$ \\
  KAIST${}^{1}$ \quad Sogang University${}^{2}$ \quad NAVER AI Lab${}^{3}$ \\
  \texttt{\{dyyjkd, haneul.yoo\}@kaist.ac.kr} \quad \texttt{hwaran.lee@sogang.ac.kr} \\}

\begin{document}

\maketitle

\begin{abstract}
    Despite the massive advancements in large language models (LLMs),  they still suffer from producing plausible but incorrect responses. 
    To improve the reliability of LLMs, recent research has focused on uncertainty quantification to predict whether a response is correct or not. 
    However, most uncertainty quantification methods have been evaluated on single-labeled questions, which removes data uncertainty—the irreducible randomness often present in user queries, which can arise from factors like multiple possible answers.  This limitation may cause uncertainty quantification results to be unreliable in practical settings. 
    In this paper, we investigate previous uncertainty quantification methods under the presence of data uncertainty. 
    Our contributions are two-fold: 1) proposing a new Multi-Answer Question Answering dataset, \data, consisting of world knowledge, mathematical reasoning, and commonsense reasoning tasks to evaluate uncertainty quantification regarding data uncertainty, and 2) assessing 5 uncertainty quantification methods of diverse white- and black-box LLMs. 
    Our findings show that previous methods relatively struggle compared to single-answer settings, though this varies depending on the task. Moreover, we observe that entropy- and consistency-based methods effectively estimate model uncertainty, even in the presence of data uncertainty. We believe these observations will guide future work on uncertainty quantification in more realistic settings. \thinspace\footnote{The code and data are available at \url{https://github.com/YangYongJin/MAQA-Official-Repo}.}
\end{abstract}
\section{Introduction}
\label{sec:intro}

\begin{figure}[t!]
    \centering
    \includegraphics[width=\columnwidth]{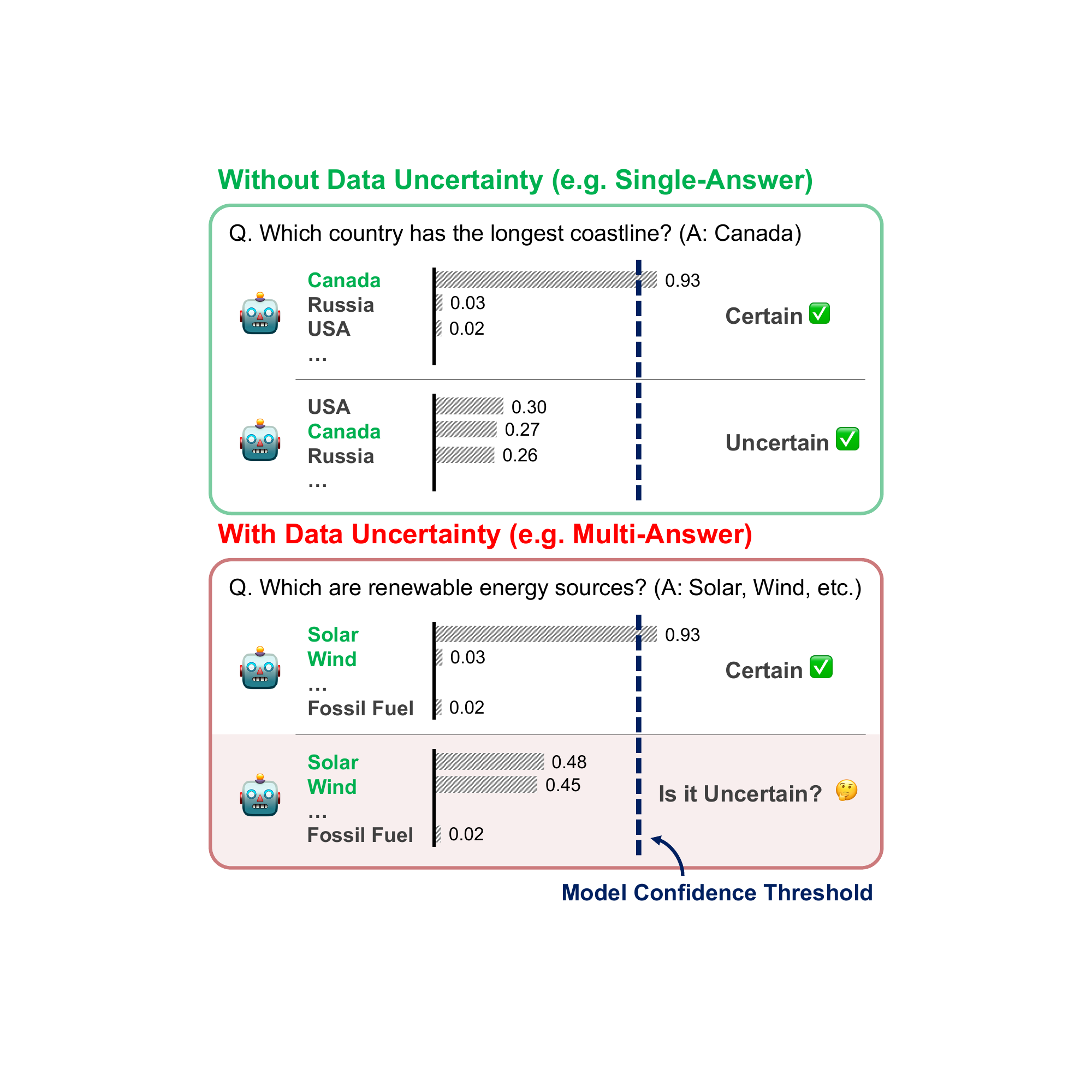}
    \caption{Evaluation settings with and without data uncertainty. When asking for a single label set, the probability distribution can be used to estimate the model uncertainty. On the other hand, when evaluating a question that has multiple answers, it may become difficult to distinguish between model uncertainty and data uncertainty, due to the existence of multiple possible answers.
    }
    \label{fig:intro_figure}
\end{figure}

Large language models~(LLMs) have demonstrated remarkable capabilities in performing diverse tasks, such as solving math problems, acquiring world knowledge, and summarizing long texts~\citep{achiam2023gpt}. 
However, these language models still suffer from hallucination~\citep{ji2023survey}, where LLMs generate false responses that appear plausible, leading users to rely on incorrect information.

To address this issue, recent studies~\citep{xiong2023can, kuhn2022semantic} 
have focused on the uncertainty quantification of LLMs, allowing the users to accept or reject responses based on the uncertainty value to improve reliability. These methods utilize either the internal states of white-box LLMs~\citep{kadavath2022language, kuhn2022semantic, yadkori2024mitigating} or response-based approaches~\citep{xiong2023can, manakul-etal-2023-selfcheckgpt} with black-box LLMs to measure the uncertainty of responses.

However, previous research mostly investigates uncertainty quantification under single-label question and answering~(QA) scenarios without considering the different sources of uncertainty: data uncertainty~(\textit{aleatoric uncertainty}) and model uncertainty~(\textit{epistemic uncertainty})~\citep{hou2023decomposing, ahdritz2024distinguishing}. Model uncertainty pertains to the insufficient capability of LLMs, indicating the model's level of knowledge about a given query. 
In contrast, data uncertainty arises from the inherent randomness in the inputs, such as when multiple answers are possible for a given query, which is the primary source of data uncertainty in our paper. 
While model uncertainty affects the reliability of LLMs, data uncertainty remains outside the model's control. 
User queries mostly introduce data uncertainty~\citep{ahdritz2024distinguishing}, making it difficult for the model to accurately estimate its own uncertainty, as shown in Figure~\ref{fig:intro_figure}.

Recently, several works~\citep{ahdritz2024distinguishing, hou2023decomposing} have attempted to estimate the uncertainty regarding data uncertainty. 
These approaches have primarily utilized ambiguous datasets~\citep{min2020ambigqa}, where each question has multiple possible answers due to its ambiguity.
To address data uncertainty, these methods employ input clarification to eliminate other answer choices, enabling them to estimate only the model uncertainty.
However, data uncertainty can still arise even when the user's intention is unambiguous, such as the question ``What are the renewable energy sources?''.
Unlike the ambiguous query, the question inherently requires multiple answers and cannot be resolved through input clarification, forcing the model to consider data uncertainty.

In this paper, we aim to enhance the current understanding of uncertainty quantification, particularly in practical scenarios where data uncertainty plays a role. To achieve this, we have two main contributions. \textit{Firstly}, we create a new benchmark, called Multi-Answer Question and Answering (\data), to introduce data uncertainty by constructing questions that clearly require a finite number of multiple answers. \textit{Secondly}, we explore uncertainty quantification methods regarding data uncertainty using the \data across various models, tasks, and quantification methods, encompassing both white-box and black-box LLMs.

Through the investigation using the \data, we found several key observations across tasks and method types. For white-box methods, data uncertainty does influence the logit distribution, causing previous uncertainty quantification methods to struggle. However, logit-based methods, especially entropy, still provide useful insights about model uncertainty, as logits tend to concentrate on a few tokens regardless of the uncertainty~($\triangleright$ Obs.~\ref{obs:white}). Also, for reasoning tasks, LLMs become overconfident after providing an initial answer, complicating uncertainty quantification~($\triangleright$ Obs.~\ref{obs:white_reasoning}). For black-box methods, although LLMs tend to overstate their confidence verbally, the consistency of their responses reliably predicts correctness across models and tasks, even regarding data uncertainty~($\triangleright$ Obs.~\ref{obs:black_verb}). Our findings suggest that uncertainty quantification methods could benefit from being developed to decompose two types of uncertainty in a task-specific manner, with a promising approach being to leverage the LLMs' probabilistic outputs.

\section{Related Work}
\label{sec:related_work}

\begin{table}[t]
\centering
\small
\resizebox{1.0\columnwidth}{!}{
\begin{tabular}{c|ccc}
\toprule[0.1em]
Dataset & Reasoning Tasks &  Answer Type & Ambiguity  \\ \midrule
Natural Questions~(\citeyear{kwiatkowski-etal-2019-natural}) & \xmark & single \& multi & \cmark\\
AmbigQA~(\citeyear{min2020ambigqa}) & \xmark & single \& multi & \cmark \\
\data (\emph{Ours}) & \cmark & multi & \xmark  \\
\bottomrule[0.1em]
\end{tabular}
}
\caption{
    Comparison of \data with previous datasets that involve some multiple-answer questions.
}
\label{tab:data_comp}
\end{table}

\subsection{Multi-Answer QA Datasets}

There are several question-answering datasets that include questions requiring multiple answers rather than a single one. 
These datasets either contain a small portion of multi-answer questions~\citep{joshi-etal-2017-triviaqa, kwiatkowski-etal-2019-natural} or include ambiguous question-answer pairs~\citep{min2020ambigqa}. 
However, most of these datasets are often confined to a single domain of world knowledge~\citep{joshi-etal-2017-triviaqa, kwiatkowski-etal-2019-natural}, with noisy labels not clarifying all the answer choices.
Moreover, in cases where multiple answers arise from ambiguity~\citep{min2020ambigqa, zhang2021situatedqa}, the question itself mostly requires a single-answer. 
Therefore, the model selects one interpretation and tries to answer with a single answer, as shown in Section~\ref{exp:additional}.
\citet{amouyal2022qampari} propose a similar type of dataset but differ in closed-book solvability and question ambiguity.
Our new benchmark includes more than 2,000 questions requiring multiple answers across tasks such as mathematical reasoning, commonsense reasoning, and world knowledge, forcing LLMs to consider data uncertainty clearly. The comparison of \data with previous datasets is shown in Table~\ref{tab:data_comp}.

\subsection{Uncertainty Quantification for LLMs}

Recently, uncertainty quantification~\citep{xiong2023can, plaut2024softmax, kadavath2022language, slobodkin-etal-2023-curious, cole2023selectively, lin2022teaching, manakul-etal-2023-selfcheckgpt, kuhn2022semantic, yadkori2024mitigating} has emerged as a significant problem in increasing the reliability of LLMs. 
There are two types of approaches for uncertainty quantification: white- and black-box based. 
White-box approaches such as max softmax logit or entropy are based on the assumption that one can access the logits~\citep{kadavath2022language, kuhn2022semantic, plaut2024softmax, yadkori2024mitigating} or hidden states of LLMs~\citep{slobodkin-etal-2023-curious, chuang2023dola}. 
In contrast, black-box based methods assume that one cannot utilize internal values, and thus use the responses of LLMs to estimate the confidence, either through verbalized confidence~\citep{lin2022teaching}, or response-consistency~\citep{xiong2023can, manakul-etal-2023-selfcheckgpt, ulmer2024calibrating}.
Based on these works, we employ both white-box and black-box approaches to analyze uncertainty quantification under data uncertainty.


\section{Multi-Answer Dataset}
\label{sec:multianswer_data}

In this section, we present our newly proposed benchmark, \data, which consists of 2,042 question-answer pairs, each requiring more than one answer. 
Our dataset includes non-ambiguous questions with clear and finite label sets and covers three distinct tasks: world knowledge, mathematical reasoning, and commonsense reasoning.

\begin{table*}[t]
\resizebox{\textwidth}{!}{
\centering
\small
\begin{tabular}{l|m{10cm}|c|c}
\toprule
Task  & Examples & \# Samples & Avg. \# Ans.\\
\midrule
World Knowledge NQ& \textbf{Question} : Who were the judges on ``Britain's Got Talent'' during its first season in 2007? & \multirow{2}{*}{592} & \multirow{2}{*}{$4.51_{\pm2.56}$}\\
(Natural Questions)& \textbf{Answer} : [
            ``Simon Cowell'',
            ``Piers Morgan'',
            ``Amanda Holden''
        ]&  & \\
\midrule
World Knowledge HLS &  \textbf{Question} : What are the members of the Commonwealth of Nations? & \multirow{2}{*}{50} & \multirow{2}{*}{{$32.67_{\pm37.47}$}}\\
(Huge Label Sets) & \textbf{Answer} : [
        ``Antigua and Barbuda'',
        ``Australia'',
        \ldots,
        ``Zambia''
    ]&  & \\
\midrule[0.05em]
\multirow{2}{*}{Mathematical Reasoning}  & \textbf{Question} : Tom plants 10 trees a year. Every year he also chops down 2 trees. He starts with 50 trees. After x years, 30\% of the trees die. Determine the possible values of x that would leave Tom with fewer than 100 trees but more than 80 trees. & \multirow{2}{*}{400} & \multirow{2}{*}{$6.86_{\pm6.29}$} \\
& \textbf{Answer} : [9, 10, 11] & &\\
\midrule[0.05em]
\multirow{2}{*}{Commonsense Reasoning} & \textbf{Question} : What are the indexes of the questions that have true answers? (a) Will Noah's Ark hypothetically sail through flooded Lincoln Tunnel? (b) Can someone with dermatitis be a hand model? \ldots (g) Could you drive a Rowe 550 to the 2008 Summer Olympics? & \multirow{2}{*}{1000} & \multirow{2}{*}{$4.04_{\pm1.47}$} \\
 &\textbf{Answer} : [
            ``d'',
            ``e'',
            ``g''
        ] & &\\

\bottomrule
\end{tabular}
}
\caption{Examples and statistics of our proposed \data.}
\label{table:data_example}
\end{table*}

\subsection{Data Collection}

We generate the \data by modifying existing benchmarks
using \texttt{GPT-4-turbo} to be multi-answer datasets as well as by creating additional question-answer sets authored by us. 
We then perform quality checks and validate the labels ourselves. The final dataset consists of 642 world knowledge question-answer pairs, 400 mathematical reasoning pairs, and 1,000 commonsense pairs, totaling 2,042 closed-book, multi-answer QA pairs. 
Detailed information on the creation of the \data, including the used prompts is provided in Appendix~\ref{appendix:dataset}.

\paragraph{World Knowledge} 

For the world knowledge evaluation set of the \data, we modify the Natural Questions~\citep{kwiatkowski-etal-2019-natural} into a multi-answer format using the LLM in four stages: filtering question-answer pairs by the number of answers, rewriting pairs with the LLM, quality checking, and human review. First, we filter the QA pairs to ensure that each question requires more than one answer. Then, we instruct the LLM to either discard pairs that ask for a single answer or refine the questions to request multiple answers. In the third stage, we conduct quality checks with the LLM to remove ambiguous questions. Finally, the authors manually review the remaining questions, ensuring the factual accuracy and validity of each answer.

Additionally, to test the behavior of LLMs with extreme cases involving a large number of answers, we generate 50 additional questions that require more than 10 answers (\ie Huge Label Sets~(HLS)). The final data consists of 592 questions generated using the NQ dataset~(World Knowledge NQ) and 50 questions that require a large number of answers (World Knowledge HLS), totaling 642 questions.

\paragraph{Mathematical Reasoning}

For the mathematical reasoning dataset, we generate 200 new questions that require multi-answer sets, covering diverse subjects such as algebra, geometry, graphs, linear algebra, and others. Additionally, we modify 150 GSM8k~\citep{cobbe2021training} questions and 50 MMLU high school questions~\citep{hendrycks2020measuring} into a multi-answer format using LLMs, through the same process as creating the dataset for the world knowledge task. Finally, the authors manually annotate all the answers for these 400 questions, each requiring multiple answers.

\paragraph{Commonsense Reasoning}

For commonsense reasoning, we modify the StrategyQA~\citep{geva2021did} dataset, which consists of true-false questions that require a reasoning process to answer the question. 
We reformulate a multi-answer question by presenting multiple true-false questions from Strategy QA and requiring the selection of questions with true answers (\ie, true statements). 
The answers are formatted as a list of question indexes.
Specifically, from the StrategyQA dataset, we randomly select questions within the range of 5-15, including at least 2 true and 2 false answers, to form a single question-answers pair. The process is repeated to create a total of 1000 pairs.

\subsection{Data Analysis}

Table~\ref{table:data_example} presents examples and statistics of the \data. As observed, the generated questions require multiple answers without ambiguity. 
The questions can ask for people, nations, numbers, or the indexes of true-false questions.

The resulting dataset has a diverse range of answers, indicated by the average number of answers for each task and the standard deviation, which varies significantly. This makes the dataset highly suitable for analyzing data uncertainty. The final dataset comprises 2,042 QA pairs, covering three different tasks with a varying number of answers for each question, with each question not ambiguous and labeled with all possible finite answers. A more detailed category distribution of the data is provided in Appendix~\ref{appendix:dataset_analysis}.



\section{Experimental Settings}
\label{sec:exp_setting}

In this section, we present experimental setups, including datasets, uncertainty quantification methods, evaluation metrics, and models that we use to explore the uncertainty quantification methods regarding data uncertainty.

\subsection{Datasets}

\paragraph{Multi-Answer Datasets}
For the evaluation under data uncertainty, we assess uncertainty quantification methods using our newly proposed \data. 
As detailed in Section~\ref{sec:multianswer_data}, this dataset consists of three different tasks: world knowledge, mathematical reasoning, and commonsense reasoning.

\paragraph{Single-Answer Datasets}

We also evaluate single QA sets with similar tasks to compare the effects of the multi-answer setting. For world knowledge, we extract questions with a single answer from the NQ-open~\citep{kwiatkowski-etal-2019-natural} dataset, totaling 1,288 pairs. For mathematical reasoning, we use the GSM8k dataset with 1,319 pairs. For commonsense reasoning, we use the StrategyQA dataset~\citep{geva2021did}, that consists of 2,290 QA pairs. Moreover, we also evaluate the uncertainty quantification methods on the mixture of single- and multi-answer datasets, denoted as \textit{all}.


\subsection{Uncertainty Quantification}

In the following section, we will introduce the white- and black-box based uncertainty quantification methods used in our analysis. 

\subsubsection{White-box LLMs} Multiple methods have been proposed to measure uncertainty using the internal states of white-box LLMs~\citep{slobodkin-etal-2023-curious,  kuhn2022semantic, plaut2024softmax}. Here, we explore the most common approaches based on the probability distribution of the next token.

\paragraph{Max Softmax Logit} Max softmax logit has been widely used for measuring the confidence or uncertainty of deep neural networks~\citep{fomicheva2020unsupervised}. For the LLMs, let $\mathbf{z} = (z_1, z_2, \ldots, z_n)$ be the logit outputs by the model before normalization, where $n$ is the vocab size. Using the maximum logit value $z_{\text{max}} = \max z_j$, the maximum softmax logit can be expressed as $\sigma(z_{\text{max}}) = \frac{e^{z_{\text{max}}}}{\sum_{j=1}^n e^{z_j}}$,
\noindent where $\sigma$ is a softmax operation. High values of $\sigma(z_{\text{max}})$ suggest the model is confident in its prediction, while lower values indicate high uncertainty. In our experiments, we use the logit values of the first token of each answer~\citep{slobodkin-etal-2023-curious}, both for single- and multi- answer datasets.

\paragraph{Entropy} Entropy is also a popular measure for estimating the uncertainty~\citep{fomicheva2020unsupervised}, which quantifies the randomness in the predicted probability distribution over the possible tokens. For a probability distribution $\mathbf{p} = (p_1, p_2, \ldots, p_n)$ of the next token, entropy $H(\mathbf{p})$ is defined as
$H(\mathbf{p}) = -\sum_{i=1}^n p_i \log p_i$,
where $p_i = \sigma(z_i)$ is the probability of the $i$-th token. High entropy values indicate that the model's predictions are spread out over many tokens, suggesting greater uncertainty. For the entropy, we use the logit values of the first token of each answer.

\paragraph{Margin} Softmax logit margin can also be used to measure the model's uncertainty~\citep{schuster2022confident}, which is defined as the difference between the largest and the second largest softmax logits. Let $p_{\text{max1}} = \max(p_1, p_2, \ldots, p_n)$ be the largest logit and $p_{\text{max2}} = \max(\{p_i\} \setminus \{p_{\text{max1}}\})$ be the second largest logit. The margin $M$ is then given by $M = p_{\text{max1}} - p_{\text{max2}}.$ A larger margin between the top tokens indicates lower uncertainty, while a smaller margin suggests higher uncertainty. 
Likewise the other methods, we use the 
logit values of the first token of each answer.

\subsubsection{Black-box LLMs}

As some proprietary models do not support logit information, uncertainty quantification using only the responses of LLMs has been well studied~\citep{manakul-etal-2023-selfcheckgpt, lin2022teaching}. In our experiments, we investigate the two most popular approaches.

\paragraph{Verbalized Confidence} The concept of verbalized confidence~\citep{lin2022teaching} involves the model explicitly stating its confidence level in its generated response. Specifically, we ask the model to provide a single answer or multiple answers~(see Appendix~\ref{appendix:implementation_details}), and then provide a confidence score, which is a numerical value in the range of 0-100. 

\paragraph{Response Consistency} Response consistency assesses uncertainty by generating multiple responses to the same prompt and analyzing the differences among them. A high degree of consistency in the responses suggests greater confidence, while diverse responses indicate higher uncertainty. Specifically, let $\{r_1, r_2, \ldots, r_m\}$ be the set of responses generated by the LLMs for a given question, where $m$ is the number of responses. The response consistency can be formulated as follows:
\vspace{-2mm}
\begin{equation*}
\text{consistency} = \frac{2}{m(m-1)} \sum_{i=1}^{m-1} \sum_{j=i+1}^{m} \text{sim}(\mathbf{r}_i, \mathbf{r}_j),
\vspace{-2mm}
\end{equation*}

\noindent where $\text{sim}$ is the function that calculates the similarity between two texts. We utilize an exact match for the similarity function.

\begin{table*}[t]
\centering
\small
\resizebox{0.95\textwidth}{!}{
\begin{tabular}{cccccccccccccc}
\toprule[0.1em]
\multirow{2}{*}{Model} & \multirow{2}{*}{Method} & \multicolumn{3}{c}{World Knowledge} & \multicolumn{3}{c}{Mathematical Reasoning} & \multicolumn{3}{c}{Commonsense Reasoning}  & \multicolumn{3}{c}{Overall}\\
\cmidrule{3-14}
& & single & multi & all & single & multi & all & single & multi & all & single & multi & all\\
\midrule
\midrule

\multirow{3}{*}{Qwen1.5-7b} & Max Logit & \textbf{73.27} & 64.68 & 67.65 & \textbf{74.36} & 64.06 & 70.70 & \textbf{59.88} & 54.05 & 57.78 & \textbf{69.17} & 60.93 & 65.38 \\
& Entropy & \textbf{71.75} & 64.35 & 66.54 & \textbf{74.15} & 64.47 & 70.72 & \textbf{60.13} & 54.10 & 58.02 & \textbf{68.68} & 60.97 & 65.09 \\
& Margin & \textbf{71.78} & 62.38 & 65.50 & \textbf{74.23} & 64.07 & 70.59 & \textbf{59.64} & 54.03 & 57.57 & \textbf{68.55} & 60.16 & 64.55 \\
\midrule
\multirow{3}{*}{Mistral-v02-7b} & Max Logit & \textbf{54.55} & 52.47 & 52.63 & \textbf{66.17} & 60.62 & 62.34 & 36.97 & \textbf{42.88} & 40.12 & \textbf{52.56} & 51.99 & 51.70 \\
& Entropy & \textbf{69.03} & 68.69 & 68.24 & \textbf{65.72} & 61.49 & 62.46 & \textbf{63.58} & 54.28 & 56.86 & \textbf{66.11} & 61.49 & 62.52 \\
& Margin & \textbf{55.06} & 52.85 & 53.11 & \textbf{66.30} & 60.60 & 62.46 & 37.85 & \textbf{43.15} & 40.58 & \textbf{53.07} & 52.20 & 52.05 \\
\midrule
\multirow{3}{*}{Llama3-8b} & Max Logit & \textbf{81.73} & 69.64 & 74.09 & \textbf{64.85} & 59.21 & 62.48 & \textbf{58.22} & 48.72 & 49.97 & \textbf{68.27} & 59.19 & 62.18 \\
& Entropy & \textbf{81.03} & 70.05 & 74.24 & \textbf{64.83} & 59.15 & 62.54 & \textbf{59.23} & 48.71 & 50.20 & \textbf{68.36} & 59.30 & 62.33 \\
& Margin & \textbf{79.51} & 67.46 & 71.61 & \textbf{64.73} & 59.01 & 62.44 & \textbf{57.54} & 48.73 & 49.80 & \textbf{67.26} & 58.40 & 61.28 \\
\midrule
\multirow{3}{*}{Mixtral-8x7b} & Max Logit & 48.82 & \textbf{53.70} & 52.93 & \textbf{56.12} & 55.86 & 55.40 & 41.67 & \textbf{42.90} & 40.53 & 48.87 & \textbf{50.82} & 49.62 \\
& Entropy & \textbf{72.82} & 69.40 & 71.76 & 55.83 & \textbf{60.39} & 56.70 & 58.58 & 55.48 & \textbf{59.39} & 62.41 & 61.76 & \textbf{62.62} \\
& Margin & 49.14 & \textbf{53.98} & 53.17 & 56.13 & \textbf{58.54} & 56.62 & \textbf{42.62} & 42.38 & 41.24 & 49.29 & \textbf{51.63} & 50.34 \\
\midrule
\multirow{3}{*}{Llama3-70b} & Max Logit & \textbf{78.27} & 66.00 & 69.78 & \textbf{61.45} & 56.19 & 57.94 & 51.42 & \textbf{55.53} & 53.55 & \textbf{63.71} & 59.24 & 60.43 \\
& Entropy & \textbf{79.47} & 66.45 & 70.46 & \textbf{60.87} & 55.43 & 57.28 & 51.64 & \textbf{55.46} & 53.61 & \textbf{63.99} & 59.11 & 60.45 \\
& Margin & \textbf{77.38} & 65.38 & 69.09 & \textbf{60.71} & 55.40 & 57.19 & 51.16 & \textbf{55.45} & 53.40 & \textbf{63.08} & 58.74 & 59.89 \\
\midrule
\multirow{3}{*}{Qwen1.5-72b} & Max Logit & \textbf{74.72} & 69.67 & 71.11 & 57.25 & \textbf{66.95} & 62.26 & \textbf{74.58} & 56.36 & 62.96 & \textbf{68.85} & 64.33 & 65.44 \\
& Entropy & \textbf{75.13} & 70.10 & 71.52 & 57.15 & \textbf{66.91} & 62.28 & \textbf{74.63} & 56.46 & 63.06 & \textbf{68.97} & 64.49 & 65.62 \\
& Margin & \textbf{73.53} & 68.60 & 70.01 & 57.06 & \textbf{66.79} & 62.22 & \textbf{74.52} & 56.30 & 62.88 & \textbf{68.37} & 63.90 & 65.04 \\
\midrule
\multirow{3}{*}{Average} & Max Logit & \textbf{68.56} & 62.69 & 64.70 & \textbf{63.37} & 60.48 & 61.85 & \textbf{53.79} & 50.08 & 50.82 & \textbf{61.91} & 57.75 & 59.12 \\
& Entropy & \textbf{74.87} & 68.17 & 70.46 & \textbf{63.09} & 61.31 & 62.00 & \textbf{61.30} & 54.08 & 56.86 & \textbf{66.42} & 61.19 & 63.10 \\
& Margin & \textbf{67.73} & 61.78 & 63.75 & \textbf{63.19} & 60.73 & 61.92 & \textbf{53.89} & 50.01 & 50.91 & \textbf{61.60} & 57.50 & 58.86 \\
\midrule
\bottomrule[0.1em]
\end{tabular}
}
\caption{AUROC scores obtained using white-box based uncertainty quantification models on different datasets and models. A high score indicates high quantification performance. ``single'' refers to the results on a QA set with only single answers, ``multi'' refers to the \data, and ``all'' refers to the combination of these two datasets, which includes both single and multi-answer sets. \textbf{Bold} denotes the row-wise maximum AUROC for each task.}
\label{tab:main_result_white}
\end{table*}
\begin{figure*}[t]
\centering
\small
    \begin{subfigure}[b]{0.32\textwidth}
    \centering
    \small
    \includegraphics[width=1.0\linewidth]{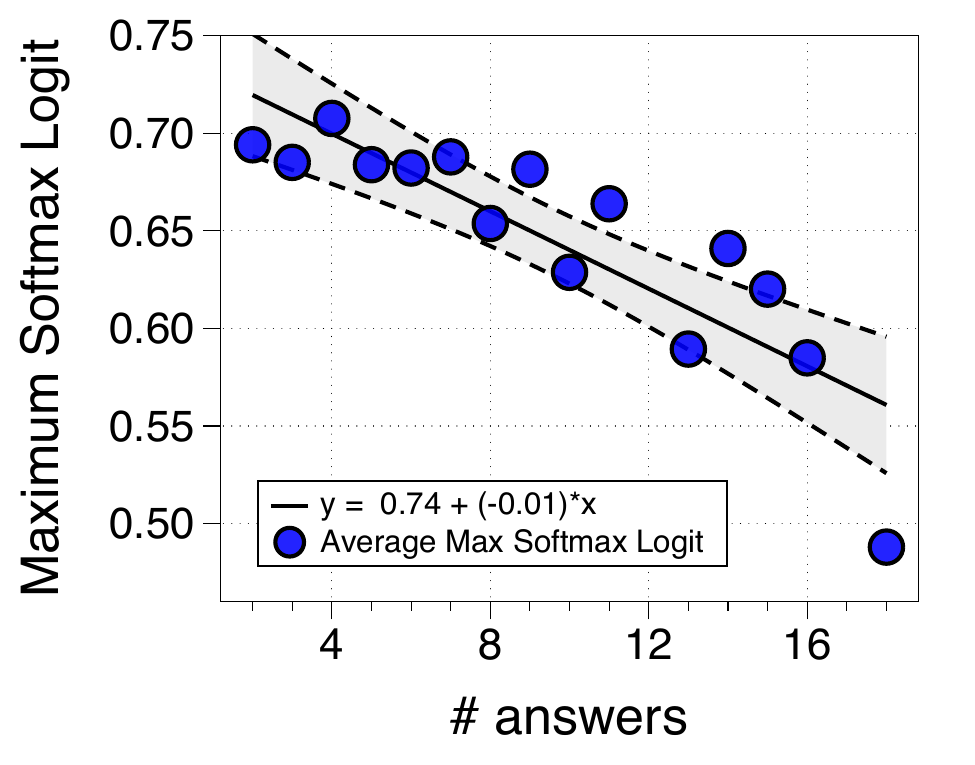}
        \caption{
        Max logit per answer count
        }\label{fig:answer_auc}
    \end{subfigure}
    \hspace{-5pt}
    \begin{subfigure}[b]{0.32\textwidth}
    \centering
    \small
    \includegraphics[width=1.0\linewidth]{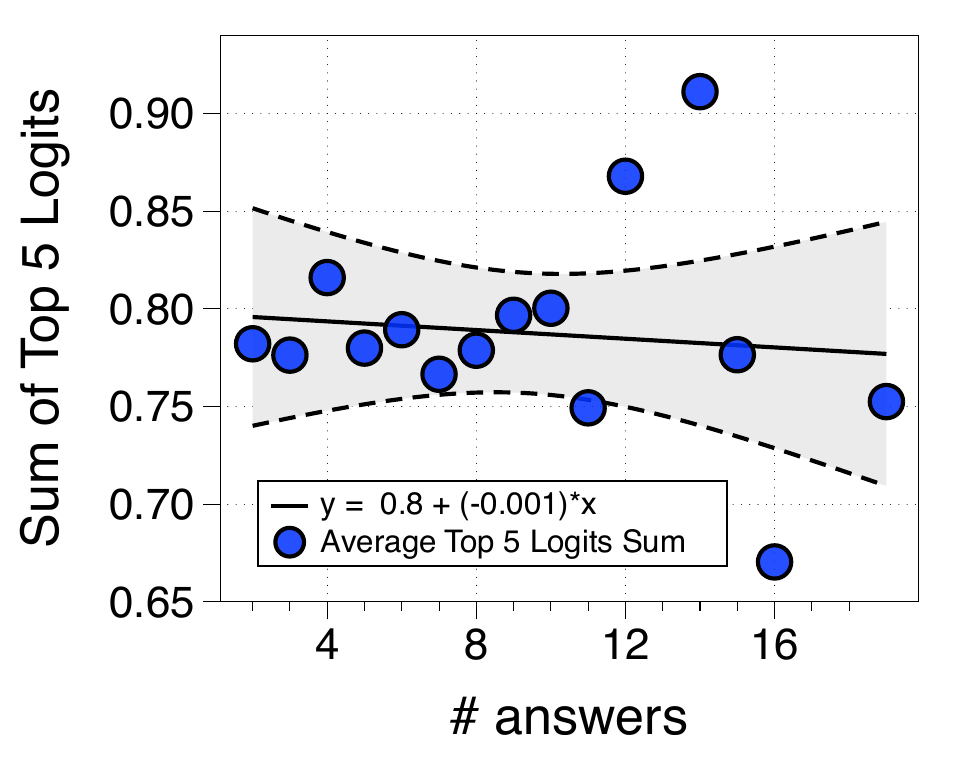}
        \caption{
        Top 5 logits sum per answer count
        }\label{fig:priority}
    \end{subfigure}
    \hspace{-5pt}
    \begin{subfigure}[b]{0.32\textwidth}
    \centering
    \small
    \includegraphics[width=1.0\linewidth]{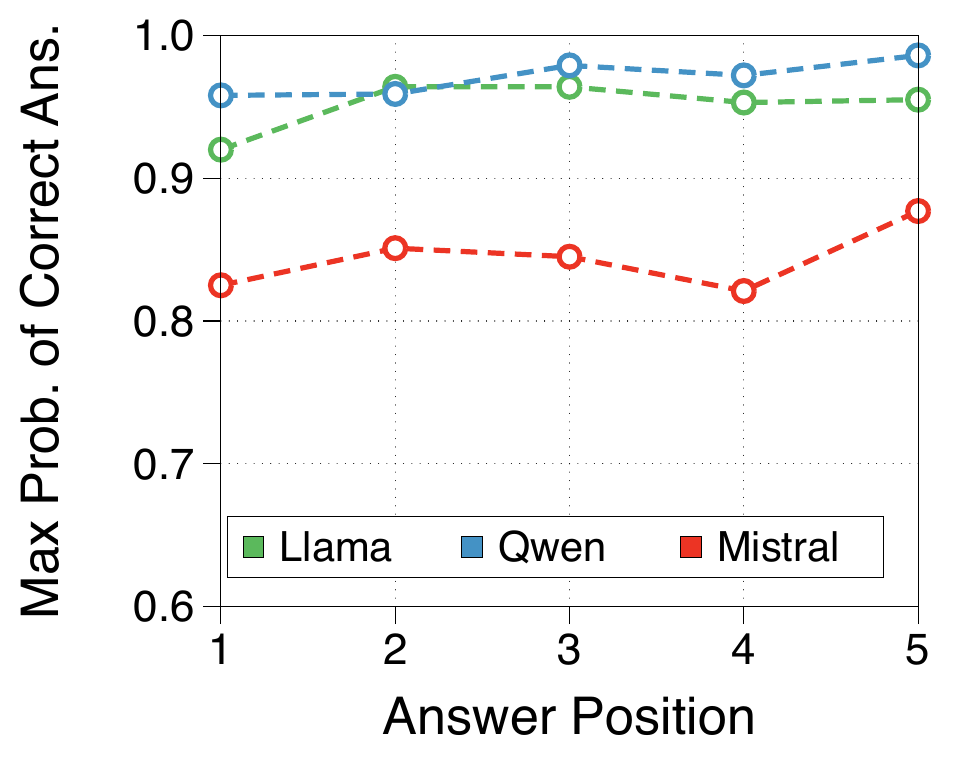}
        \caption{
        Max logit of each answer position
        }\label{fig:seq_math}
    \end{subfigure}
    \caption{(a) Maximum probability of correct answer by the number of answers when evaluated on world knowledge part of \data. The number of answers clearly affects the probability value, indicating the data uncertainty. (b) Sum of top 5 probabilities of correct answer per each number of answers, which seems constant across answer count. The results are averaged over three 7-8B models. (c) Max probabilities of correct answers per each answer position when evaluated on reasoning tasks. LLMs tend to be overconfident, especially after the first answer. }
\end{figure*}

\subsection{Evaluation Metrics}

\paragraph{Metrics for Correctness}

To calculate the correctness of a single answer, we use the \textit{accuracy} with the exact match between the predicted answer and the ground-truth answer. For the correctness of multiple answers, we adopt three metrics. The main metric, as we aim to assess the reliability of each answer in the response, is \textit{precision}. This metric calculates the proportion of correctly predicted answers out of all predicted answers, using an exact match. We also define \textit{recall} and \textit{F1 score} in our setting, with details in Appendix~\ref{appendix:implementation_details} , and results using these additional metrics in Appendix~\ref{appendix:more_results}.


\paragraph{Metrics for Uncertainty Quantification}

High performance in uncertainty quantification indicates that the uncertainty measure can effectively predict whether the model's predictions are likely to be correct or incorrect. Note that our evaluation does not focus on calibration, which aims to predict the exact correctness score using the confidence value. We primarily use the Area Under the Receiver Operating Characteristic Curve (AUROC) for failure prediction, which provides a comprehensive evaluation of the model's ability to distinguish between correct and incorrect predictions. Results using Area under the Precision-Recall Curve (AUPRC) are presented in Appendix~\ref{appendix:more_results}.

\subsection{Evaluation Models and Inference}

We use Llama3-(8b, 72b)~\citep{llama3modelcard}, Qwen1.5-(7b, 72b)~\citep{bai2023qwen},
Mistral-v02-7b~\citep{jiang2023mistral}, and Mixtral-8x7b~\citep{jiang2024mixtral} for white-box LLMs.
For black-box LLMs, we use GPT-3.5-turbo-0125~\citep{ouyang2022training} and GPT-4-turbo-0125~\citep{achiam2023gpt}, along with some white-box LLMs used as black-box LLMs. For the evaluation of white-box LLMs, we use greedy sampling with a temperature value of 1.0. For the black-box LLMs, we sample 5 responses for each question, using a temperature value of 0.99 and top-p sampling with p equal to 0.9. We use vanilla prompting for world knowledge questions, where the instruction in the prompt guides the model to answer in a specific format.
On the other hand, we utilize the Chain of Thought~(CoT) prompting~\citep{wei2022chain} for reasoning tasks, as vanilla prompting significantly degrades performance. 
More implementation details are presented in Appendix~\ref{appendix:implementation_details}.

\section{Experimental Results}
\label{sec:results}

\begin{table*}[tp]
\centering
\small
\resizebox{0.95\textwidth}{!}{
\begin{tabular}{cccccccccccccc}
\toprule
\multirow{2}{*}{Model} & \multirow{2}{*}{Method} & \multicolumn{3}{c}{World Knowledge} & \multicolumn{3}{c}{Mathematical Reasoning} & \multicolumn{3}{c}{Commonsense Reasoning}  & \multicolumn{3}{c}{Overall} \\
\cmidrule{3-14}
& & single & multi & all & single & multi & all & single & multi & all & single & multi & all\\ 
\midrule
\midrule

\multirow{2}{*}{Qwen1.5-7b} & Verbalize & \textbf{56.36} & 54.10 & 53.67 & \textbf{69.43} & 58.71 & 65.40 & \textbf{58.16} & 56.74 & 52.49 & \textbf{61.32} & 56.52 & 57.19 \\
 & Consistency & \textbf{81.56} & 62.11 & 81.07 & \textbf{94.58} & 91.35 & 94.23 & 60.78 & \textbf{61.17} & 60.78 & \textbf{78.97} & 71.54 & 78.69 \\
\midrule
\multirow{2}{*}{Mistral-v02-7b} & Verbalize & 65.82 & 66.21 & \textbf{68.57} & 57.78 & \textbf{64.82} & 58.90 & \textbf{62.17} & 47.39 & 51.84 & \textbf{61.92} & 59.47 & 59.77 \\
 & Consistency & 75.83 & \textbf{85.02} & 79.53 & 94.55 & 94.57 & \textbf{94.63} & 60.20 & \textbf{68.26} & 63.75 & 76.86 & \textbf{82.62} & 79.30 \\
\midrule
\multirow{2}{*}{Llama3-8b} & Verbalize & \textbf{71.68} & 64.24 & 69.54 & 61.02 & 60.51 & \textbf{62.39} & \textbf{59.56} & 50.10 & 59.15 & \textbf{64.09} & 58.28 & 63.69 \\
 & Consistency & \textbf{87.51} & 85.20 & 86.89 & 92.59 & 91.24 & \textbf{92.87} & \textbf{65.74} & 62.97 & 55.27 & \textbf{78.62} & 77.14 & 75.01 \\
\midrule
\multirow{2}{*}{GPT-3.5} & Verbalize & 62.68 & 60.33 & \textbf{64.98} & 58.34 & \textbf{61.30} & 50.29 & 55.59 & \textbf{59.46} & 58.04 & 58.87 & \textbf{60.36} & 57.77 \\
 & Consistency & 80.93 & \textbf{84.31} & 81.53 & 87.90 & 87.07 & \textbf{88.04} & 63.61 & \textbf{64.94} & 63.00 & 77.48 & \textbf{78.77} & 77.52 \\
\midrule
\multirow{2}{*}{GPT-4} & Verbalize & \textbf{69.28} & 64.76 & 68.42 & 72.23 & \textbf{72.45} & 68.58 & \textbf{55.80} & 50.11 & 52.13 & \textbf{65.77} & 62.44 & 63.04 \\
 & Consistency & \textbf{78.92} & 77.44 & 71.30 & \textbf{93.85} & 92.15 & 88.04 & \textbf{72.14} & 71.54 & 60.65 & \textbf{81.64} & 80.38 & 73.33 \\
\midrule
\multirow{2}{*}{Average} & Verbalize & \textbf{65.16} & 61.93 & 65.04 & \textbf{63.76} & 63.56 & 61.11 & \textbf{58.26} & 52.76 & 54.73 & \textbf{62.39} & 59.42 & 60.29 \\
 & Consistency & \textbf{80.95} & 78.82 & 80.06 & \textbf{92.69} & 91.28 & 91.56 & 64.49 & \textbf{65.78} & 60.69 & \textbf{78.71} & 78.09 & 76.77 \\
\midrule
\bottomrule
\end{tabular}
}
\caption{The AUROC scores obtained using black-box based uncertainty quantification models on different datasets and models. A high score indicates high uncertainty quantification performance. ``single'' refers to the results on a QA set with only single answers, ``multi'' refers to the \data, and ``all'' refers to the mixture of these two datasets, which includes both single and multi-answer sets. \textbf{Bold} denotes the row-wise maximum AUROC for each task.}
\label{tab:main_result_black}
\end{table*}

\subsection{Results of White-box based UQ methods}

Table~\ref{tab:main_result_white} shows the AUROC scores of different uncertainty quantification methods for white-box LLMs across various tasks and models. Based on the results, we have two key observations.
 
\begin{observation}\label{obs:white}
     Data uncertainty does impact logit distributions in the world knowledge task; however, the logits---particularly those represented as entropy---remain useful for predicting the factuality of responses due to internal prioritization. 
\end{observation}

In the world knowledge task, we observe that the AUROC scores for both the \textit{multi} and \textit{all} decline across almost all models and methods compared to the single-answer dataset evaluation. This decrease can be attributed to the data uncertainty, as previous methods do not account for the composition of data and model uncertainties separately.
Figure~\ref{fig:answer_auc} supports this claim, showing that the maximum logit values of the correct answers decrease as the number of ground-truth answers increases.

However, despite the impact of data uncertainty, logit values, especially entropy, remain useful for predicting the correctness of answers, as evidenced by the average AUROC higher than 60.
The effectiveness of entropy indicates that LLMs may prioritize a few tokens when generating each answer regardless of the number of answers.

As depicted in Figure~\ref{fig:priority}, the sum of the top 5 softmax logits of the correct answers remains consistent regardless of the number of ground-truth answers.
Moreover, when LLMs are asked to provide multiple answers in an order that seems more probable and common, the AUROC scores increase, especially when the true answer space is larger, as detailed in Appendix~\ref{appendix:more_results}. 
This supports the claim that LLMs have their own internal priority of answers, making the model less affected by data uncertainty.

\begin{figure*}[t]
\centering
\small
    \begin{subfigure}[b]{0.32\textwidth}
    \centering
    \small
    \includegraphics[width=1.0\linewidth]{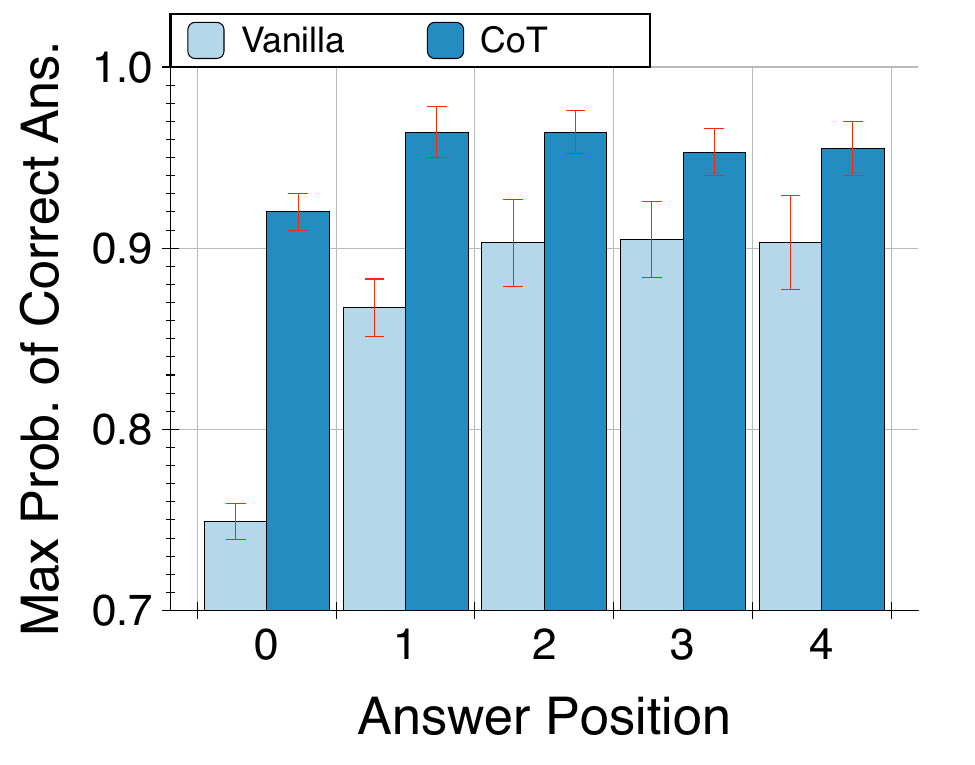}
        \caption{
        Result using different promptings
        }\label{fig:reasoning_why}
    \end{subfigure}
    \hspace{-5pt}
    \begin{subfigure}[b]{0.32\textwidth}
    \centering
    \small
    \includegraphics[width=1.0\linewidth]{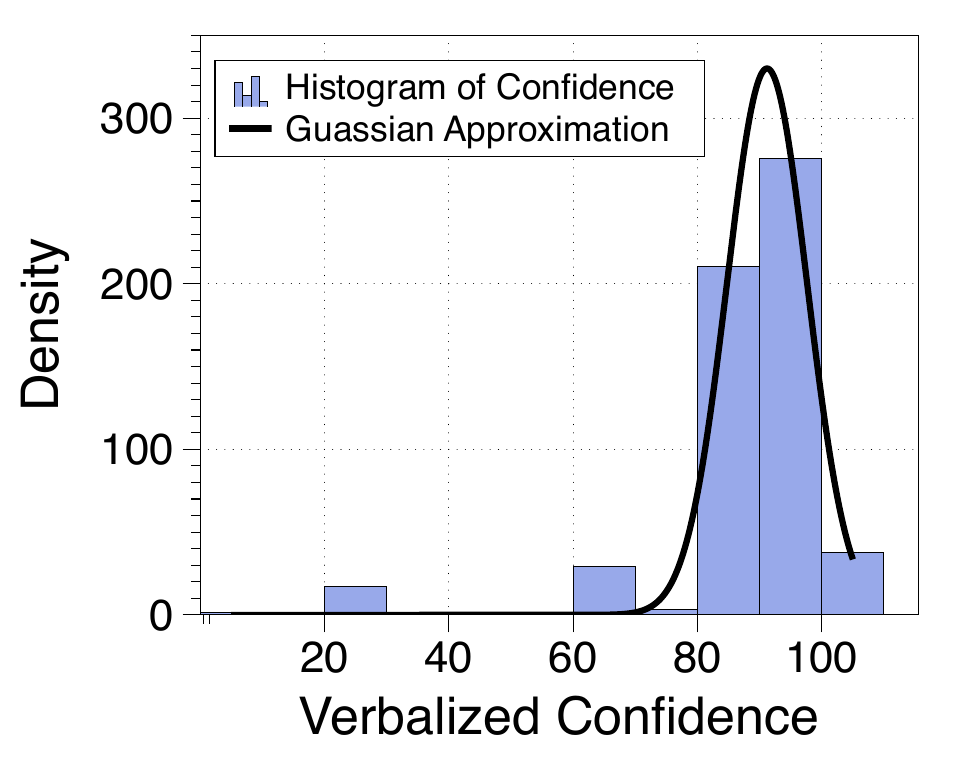}
        \caption{
        Histogram of verbalized confidence 
        }\label{fig:overconfidence}
    \end{subfigure}
    \hspace{-5pt}
    \begin{subfigure}[b]{0.32\textwidth}
    \centering
    \small
    \includegraphics[width=1.0\linewidth]{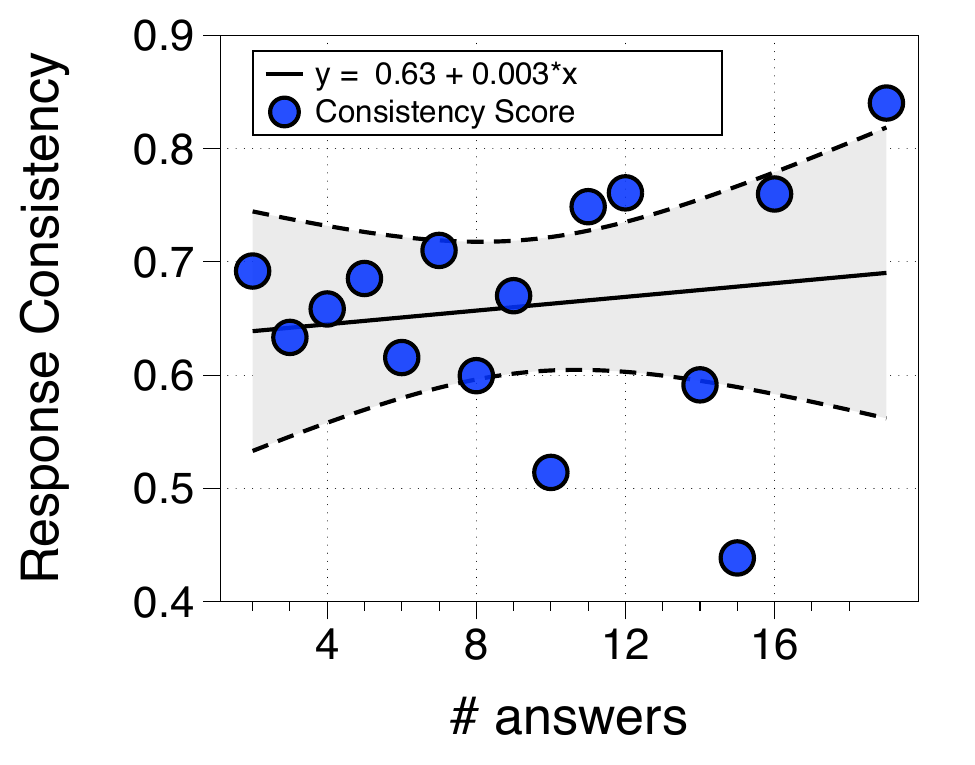}
        \caption{
        Consistency per answer count
        }\label{fig:consistency}
    \end{subfigure}
    \caption{(a) Maximum probability of correct answer per answer position by different prompting methods. CoT prompting clearly increases the confidence score. (b) Histogram of verbalized confidence values when evaluated on \data. LLMs tend to be overconfident, as confidence score is concentrated in the range of 80-100. (c) Response consistency per answer count. Score is averaged over all datasets. For all three results, Llama-3-8b is utilized. }
\end{figure*}

\begin{observation}\label{obs:white_reasoning}
      In reasoning tasks, LLMs tend to be overconfident even under data uncertainty, especially after providing the first answer, complicating the performance of uncertainty quantification.
\end{observation}

In the mathematical and commonsense reasoning tasks, where LLMs need to output answers after multiple intermediate reasoning steps, we observe that the performance of uncertainty quantification based on logit values also decreases. This trend becomes more pronounced in the multi-answer case, as performance mostly decreases compared to single-answer cases.

This can be explained by the LLMs' tendency to be overconfident on reasoning tasks. As illustrated in Figure~\ref{fig:seq_math}, the average maximum softmax logit for each answer token tends to be high, exceeding 0.9 in some models (e.g., Llama3, Qwen1.5). This value increases as the LLMs continue generating subsequent answers. This suggests that LLMs become progressively more overconfident after producing the initial answer with the reasoning process, complicating uncertainty quantification, especially in scenarios involving multiple answers. 

To support this observation, we compare the results of CoT with vanilla prompting as depicted in Figure~\ref{fig:reasoning_why}.
The results show that CoT prompts significantly increase the logit values, indicating that the reasoning process further boosts the LLMs' confidence.

\subsection{Results of Black-box based UQ methods}

Table~\ref{tab:main_result_black} shows the results of uncertainty quantification methods for black-box LLMs, resulting in a key observation.

\begin{observation}\label{obs:black_verb}
     LLMs tend to be overconfident when verbally providing a confidence score regardless of the data uncertainty, making uncertainty quantification difficult. On the other hand, utilizing response consistency works extremely well on all tasks, regardless of single- or multi-answers.
\end{observation}

As shown in Table~\ref{tab:main_result_black}, verbalized confidence struggles, as evidenced by the relatively low AUROC scores, especially on the commonsense reasoning task. The trend is similar for both single-answer and multi-answer datasets, as verbalized confidence is usually less affected by data uncertainty since the confidence is measured by LLMs themselves. The low AUROC score of verbalized confidence can be explained by overconfidence. As shown in Figure~\ref{fig:overconfidence}, for both single-answer and multi-answer cases, the confidence is mostly distributed above 80 on a scale of 0-100. This overconfidence of LLMs aligns with the findings of \citet{xiong2023can} for single-answer cases and is also evident for multi-answer cases.


In contrast, consistency-based methods have significantly high AUROC scores on all tasks, especially on multi-answer datasets. This implies that even with multiple possible answers, the consistency is strongly correlated with model correctness. Figure~\ref{fig:consistency} demonstrates that consistency scores are less influenced by the number of true labels. Additionally, because the multi-answer approach measures finer-grained similarity based on the overlap rate across all answers, it proves highly effective for multi-answer datasets.

\subsection{Discussions}
\label{exp:additional}

\paragraph{Unifying Observations from White- and Black-Box Approaches}

Overall, the presence of data uncertainty in multi-answer sets leads to a decrease in traditional uncertainty measures, except for response consistency. Furthermore, the trend varies by task, particularly in cases where reasoning is crucial. These findings suggest that new quantification methods are needed to decompose model uncertainty without requiring input clarification, and the approach to uncertainty quantification should be tailored to the specific task at hand.

Among the evaluated uncertainty quantification methods, response consistency performs best, demonstrating the value of using multiple responses per prompt, despite the relatively high computational cost of sampling. Among white-box methods, entropy, which incorporates logit values for all vocabulary terms, either matches or outperforms other methods. This highlights the effectiveness of leveraging the model’s probabilistic outputs, whether through entropy or response variability.
Therefore, a promising approach could involve a method that takes into account this probabilistic nature, while further improvements are needed to reduce the high cost of utilizing it.

\paragraph{Performance by Model Types} As shown in Table~\ref{tab:main_result_white} and \ref{tab:main_result_black}, Llama3 and Qwen1.5 exhibit similar trends, with relatively strong quantification performance for both white- and black-box methods. In contrast, the Mistral models tend to struggle, particularly with white-box methods at a temperature value of 1.0. This may be attributed to differences in logit scales that impact the overall scale of softmax values. These issues could potentially be mitigated by using more logits, such as entropy.

\paragraph{Comparison with Performance on AmbigQA}


\begin{table}[t]
\centering
\resizebox{0.9\linewidth}{!}{
\begin{tabular}{c|ccc|ccc}
\toprule[0.1em]
\multirow{2}{*}{Method} & \multicolumn{3}{c|}{World Knowledge} & \multicolumn{3}{c}{AmbigQA} \\
\cmidrule{2-7}
& single & multi & all & single & multi & all\\
\midrule
\midrule

Max Logit & \textbf{81.73} & 69.64 & 74.09 & \textbf{79.23} & 77.22 & 77.60 \\ 
\midrule
Entropy & \textbf{81.03} & 70.05 & 74.24 & \textbf{78.55} & 76.21 & 77.73  \\
\midrule
Margin & \textbf{79.51} & 67.46 & 71.61 & \textbf{77.16} & 74.92 & 75.38   \\
\midrule
Verbalize & \textbf{71.68} & 64.24 & 69.54 & 62.34 & \textbf{64.71} & 63.16   \\
\midrule
Consistency & \textbf{87.51} & 85.20 & 86.89 & 82.26 & \textbf{82.45} & 81.82   \\

\bottomrule[0.1em]
\end{tabular}
}
\caption{The AUROC scores of different uncertainty quantification meathods using the Llama-3-8b model on two MAQA world knowledge and AmbigQA. }
\label{tab:comp_ambig}
\end{table}

To test whether the ambiguity of questions affects the performance of uncertainty quantification, we evaluate the uncertainty quantification performance using the test set from AmbigQA~\citep{min2020ambigqa}. For the AmbigQA dataset, we utilize the portion annotated with ``singleQA'' for the \textit{single} set, ``multipleQA'' for the \textit{multi} set, and a mixture of both for the \textit{all} set. Since the dataset mostly involves world knowledge, we compare the results with those from the world knowledge portion of \data. 

As shown in Table~\ref{tab:comp_ambig}, the decrease in AUROC scores when transitioning from single-answer to multi-answer or all-answer settings is significantly greater for the MAQA world knowledge dataset than for the AmbigQA dataset. This suggests that previous quantification methods are less impacted by data uncertainty arising from ambiguous questions, as most questions themselves inherently require a single answer. These differences arise from the model's behavior in response to different question types, as further explained in the qualitative comparison of model behaviors in Appendix~\ref{appendix:qual_compare_ambig}.


\section{Conclusion}
\label{sec:conclusion}


In this paper, we contribute to the uncertainty quantification of LLMs in two aspects: First, we propose a new benchmark, \data, which consists of question-answer pairs where each question requires more than two unambiguous answers, ensuring data uncertainty at the question level. Second, we investigate uncertainty quantification for both white- and black-box LLMs regarding data uncertainty, finding key observations. We hope our work serves as a foundation for future research on realistic settings for uncertainty quantification.

\section*{Limitations}

We created a novel multi-answer dataset, \data, that covers three different tasks. 
Although we conducted a quality check, there may remain some ambiguous questions or unclear answers. Additionally, likewise many other QA datasets, our data contains answers that can take multiple forms. Despite our efforts to include all possible answers, there may be some noise in the dataset.

Moreover, to ensure that the answer space is precise and unambiguous, our dataset primarily consists of short-form answers. For long-form answers, it is difficult to analyze the influence of data uncertainty, and the correctness of such answers can be vague, making the assessment of uncertainty quantification performance challenging. We may extend this analysis to long-form answers in future research.

In this paper, we evaluate multiple uncertainty quantification methods for both white-box and black-box LLMs in the presence of data uncertainty. Although we have several observations, none of the methods are free of hyperparameters, such as temperature, sampling methods, etc. We believe that future work should investigate these settings further and establish guidelines for their use.

\section*{Ethics Statement}

Hallucination, where large language models (LLMs) generate responses that appear plausible but are factually incorrect, poses a significant ethical issue. This phenomenon can lead to the dissemination of misinformation, which may cause harm by misleading users, decreasing the reliability of LLMs, and potentially influencing decision-making processes in critical areas such as healthcare, legal, and financial services.  Therefore, addressing hallucinations in LLMs is crucial to ensure that AI systems operate within ethical boundaries.

Uncertainty quantification methods offer a promising approach to addressing the ethical challenges posed by hallucinations in LLMs. By estimating the confidence levels of the models' outputs, these methods can help identify and flag potentially unreliable or erroneous information. This transparency enables users to better assess the trustworthiness of AI-generated content and make informed decisions. Moreover, incorporating uncertainty quantification can guide developers in refining LLMs to reduce hallucinations, thereby enhancing the ethical deployment of AI technologies. 

\section*{Acknowledgements}

We would like to thank Namgyu Ho for his feedback on the outline of the paper. We also extend our gratitude to Jimin Lee for extensive discussions on dataset annotation.

\bibliography{custom}

\appendix

\clearpage
\section*{Appendix}

\section{Dataset Construction}
\label{appendix:dataset}

In this section, we will explain in detail how each task of \data was created. We utilize the Natural Questions~\citep{kwiatkowski-etal-2019-natural}, which is under the Apache 2.0 license, as well as GSM8k~\citep{cobbe2021training}, MMLU~\citep{hendrycks2020measuring}, and StrategyQA~\citep{geva2021did}, which are under the MIT license. Our dataset \data will be distributed under the Apache 2.0 license.

\subsection{World Knowledge}

\paragraph{World Knowledge NQ}
As explained in Section~\ref{sec:multianswer_data}, we modify the question-answering pairs so that each question requires multiple answers from the Natural Questions~\citep{kwiatkowski-etal-2019-natural} dataset using the OpenAI \texttt{GPT-4-turbo} model. We adopt the following prompt template to modify the dataset (Note: some parts are skipped and marked as \ldots, as the original prompt is too long).

\begin{footnotesize}
\begin{tcolorbox}[breakable, enhanced, top=1pt, left=1pt, right=1pt, bottom=1pt]
\textbf{Prompt Template for Modifying Natural Question Dataset} \newline\newline
You are given a question and answer pair. For each question, there are multiple answers. You have two options: 1) reject a pair, or 2) refine a pair. \\
\\
Reject a pair with the answer ``reject" if the question and answer pair contains the following features:\\
\\
- The user's intention in asking the question is to receive a single answer. Therefore, if multiple answers imply the same meaning, or if there are multiple answers because the question is "ambiguous," you should reject this pair.\\
\\
    Example 1: All the answers convey the same meaning.
    ...\\
    \\
- If there are conflicts between the answers.\\
    Example: The question below has conflicting answers.
    ...\\
    \\
- If the question is time-dependent, meaning its answer can change in the future.\\
...\\    
Otherwise, refine the question and answer pair and return them. Instructions for refining are as follows:\\
\\
- Properly format the question to make it a complete sentence. Modify the question so that it clearly requires multiple different answers.\\
    Example: \\
    ... \\
    \\
- Remove all incorrect answers ...\\
    ...\\
- Remove duplicated responses ("the queen" and "The Queen" have the same meaning).\\
    ...\\
    \\
**Now, here is the question-answer pair:**\\
\\
Question : \\
\{question\} \\
Answer : \\
\{answer\} \\
\end{tcolorbox}
\end{footnotesize}

After that, we conduct an additional quality check through the LLM, assessing the validity and ambiguity of the answers, the quality of the questions, and their time-dependency, using the instruction below~\citep{zheng2024judging}:

\begin{footnotesize}
\begin{tcolorbox}[breakable, enhanced, top=1pt, left=1pt, right=1pt, bottom=1pt]
\textbf{Prompt Template for Quality Check} \newline\newline
Please act as an impartial judge and assess the quality of given question-and-answer pairs. The ideal question should naturally encourage a range of answers, rather than bundling multiple distinct questions together. Use the following criteria to determine if a pair scores highly:\\\\
1. (3 points) The question genuinely requires multiple, semantically unique responses, and should not consist of two sub-questions (e.g. What are A and B?). Also, the questions should reflect those that real users actually want to ask.\\
2. (2 points) All provided answers must be semantically distinct from one another, and one answer should not encompass another (e.g. low score cases: [ 1980, 06.1980 ], [ Vancouver, Vancouver Canada ]).\\
3.(3 points) Each answer in the list must appropriately address the question, without any missing answers or wrong answers. Missing or wrong answers would result in a low score.\\
4. (1 points) The question is unambiguous, clear and interpretable in only one way.\\
5. (1 points) The question should not be time-dependent (answers change over time).\\
\\
Begin your evaluation by providing a short and brief explanation that consist of two to three sentences. Be as objective as possible. After providing your explanation, please rate the response on a scale of 1 to 10 by strictly following this format: "[[rating]]", for example: "Rating: [[5]]". \\
\\
**Now, here is the question-answer pair:**\\
\\
Question : \\
\{question\} \\
Answer : \\
\{answer\} \\
\end{tcolorbox}
\end{footnotesize}

Based on the scores, we remove the answers that have a score lower than 5. Finally, three authors manually check the factuality, the ambiguity of the question, whether the question consists of multiple sub-questions, and also label the question type for each question. We adopted consensus validation, where all all three authors must arrive at the same answer set and agree on the question's difficulty level. The final data totals 592 pairs, covering diverse subjects and question types.

\paragraph{World Knowledge Huge}

Additionally, we generate 50 questions that require large sets of answers. Specifically, we look for Wikipedia lists that have more than 10 items for specific subjects. Using these lists as sources, we generate new questions that have all the list components as answers. The final data consists of 50 question-answer pairs that cover diverse domains. 

\subsection{Mathematical Reasoning}

\paragraph{Manual Generation}

For 200 math questions that require multiple answers, covering algebra, graphs, linear algebra, arithmetic, and other topics, we mostly set the range, such as finding the $x$ that satisfies the condition. Each condition is related to different domains, such as arithmetic, numbers, graphs, etc.

\paragraph{Modify dataset using LLMs} We also modify some GSM8k~\citep{cobbe2021training} and MMLU questions~\citep{hendrycks2020measuring} into multi-answer format.

\begin{footnotesize}
\begin{tcolorbox}[breakable, enhanced, top=1pt, left=1pt, right=1pt, bottom=1pt]
You're given a question that involves mathematical reasoning from the previous dataset. Your task is to refine this question and create a new question-answer pair. The refined question should be designed to require multiple answers, hence the answer should be presented as a list containing at least three elements. Each refined question must demand a deeper level of thought and involve complex problem-solving skills that are not trivial. Here are illustrative examples across various areas of mathematics for guidance:\\

1. **Example 1**\\
    - **Original Question:** "Jane's quiz scores were 98, 97, 92, 85, and 93. What was her mean score?"\\
    - **Refined Question:** "Jane's quiz scores were 98, 97, 92, 85, and 93. List the integer numbers that are higher than her mean score, but lower than 100."\\
    - **Answer:** [94, 95, 96, 97, 98, 99]\\
2. **Example 2**\\
    - **Original Question:** "What is the second number in the row of Pascal's triangle that has 43 numbers?"\\
    - **Refined Question:** "List the unique numbers in the row of Pascal's triangle that has 6 numbers."\\
    - **Answer:** [1, 5, 10]\\
3. **Example 3**\\
    - **Original Question:** "How many arithmetic sequences of consecutive odd integers sum to 240?"\\
    - **Refined Question:** "List the smallest number in of arithmetic sequences that contain consecutive odd integers that sum to 240."\\
    - **Answer:** [9, 15, 23, 35, 57, 119]\\

If the given question seems too challenging to refine, you may generate a new, simpler question that necessitates more than four answers.\\

Now here is the question that you need to refine\\
\end{tcolorbox}
\end{footnotesize}

\subsection{Commonsense Reasoning}

For commonsense reasoning, we modify the StrategyQA~\citep{geva2021did} dataset, which includes multiple true-false questions that require a reasoning process to answer.
To create a question that needs multiple answers from each true-false question, we design a task that asks for the indexes of all questions with true answers~(true statements), given multiple questions with answers that are either true or false.
Specifically, from the StrategyQA dataset, we randomly select questions ranging from 5-15 so as not to exceed the maximum input prompt length, ensuring at least 2 true and 2 false statements in each selection. This process is repeated until we generate 1000 question-answer pairs.

\section{Dataset Distribution}
\label{appendix:dataset_analysis}

\begin{table}[t]
\centering
\resizebox{\linewidth}{!}{
\begin{tabular}{@{}lr|lr@{}}
\toprule
\textbf{Task}                      & \textbf{\#}          & \textbf{Category}                                           & \textbf{\#} \\ \midrule

\multirow{11}{*}{World Knowledge NQ}         & \multirow{11}{*}{592} & History                                      & 103 \\
                                       &                      & Sports                   & 72 \\
                                       &                      & Geography                                & 58 \\
                                       &                      & Movie                                      & 57 \\
                                       &                      & Science                             & 56 \\
                                       &                      & Literature                                                     & 41 \\
                                       &                      & Entertainment                                                     & 34 \\
                                       &                      & Politics                                                     & 27\\
                                       &                      & Music                                                     & 26\\
                                       &                      & Art                                                     & 10\\
                                       &                      & Others (Religion, Education, etc)                                                     & 108\\ \midrule
\multirow{5}{*}{World Knowledge HLS}     & \multirow{5}{*}{50} & Science                                                       & 17 \\
                                       &                      & Geography                                       & 14 \\
                                       &                      & History                      & 9 \\
                                       &                      & Entertainment                                          & 6 \\
                                       &                      & Culture                                               & 5  \\
                                       \midrule

\multirow{6}{*}{Mathematical Reasoning}     & \multirow{6}{*}{400} & Arithmetic and Number Theory                                                       & 215 \\
                                       &                      & Algebra                                       & 118 \\
                                       &                      & Statistics                      & 26 \\
                                       &                      & Combinatorics and Probability                                          & 17 \\
                                       &                      & Graph Theory                                               & 14 \\ &
                                       & Geometry                                               & 10 \\
                                       \midrule
\bottomrule
\end{tabular}}
\caption{Detailed statistics of each dataset, including the number of question-answer pairs for each category of each task. We omit the statistics of commonsense reasoning as it is originated from the single StrategyQA~\citep{geva2021did} dataset.}
\label{tab:data_stats}
\end{table}

Table~\ref{tab:data_stats} shows the detailed statistics of our proposed \data benchmark. As observed, our benchmark covers a wide range of categories for world knowledge, including history, sports, geography, and others. Additionally, we cover multiple problems for mathematical reasoning, involving arithmetic, algebra, graph theory, and others, making \data a strong benchmark that can test a wide range of domains and tasks.


\section{Implementation Details}
\label{appendix:implementation_details}

\paragraph{Inference}

For white-box LLM experiments, we use greedy sampling with a temperature value of 1.0 to get the normalized probability logit. For all models, we use float 16 precision to save memory. All our experiments are conducted on 4 NVIDIA A100 GPUs. For the black-box LLMs, we utilize white-box LLMs such as Llama3-8b, Qwen1.5-7b, and Mistral-v02, and additionally OpenAI GPT-3.5 and GPT-4 using the OpenAI API. We adopt top-p sampling with p equal to 0.9.

For the prompting method, we use vanilla prompting for world knowledge, as shown below:

\begin{footnotesize}
\begin{tcolorbox}[breakable, enhanced, top=1pt, left=1pt, right=1pt, bottom=1pt]
Instruction : Given a question that has multiple answers, answer the question following the instructions below:\\
\\
1. Keep your response as brief as possible without any explanation.\\
2. Mark each answer with a number followed by a period.\\
3. Separate each answer with a number, a comma and a space.\\
\\
The format of the answer should be given as follows:\\
\\
1.YourAnswer1, 2.YourAnswer2, 3.YourAnswer3\\
\\
Now, please answer this question.\\
\\
Input : \\
\\
Question: \\
\\
\{ question \} \\
\end{tcolorbox}
\end{footnotesize}

For reasoning tasks, we employ CoT~\citep{wei2022chain} prompting. To make LLMs do CoT reasonings, we add the instruction of ``explain step-by-step'' as follows:

\begin{footnotesize}
\begin{tcolorbox}[breakable, enhanced, top=1pt, left=1pt, right=1pt, bottom=1pt]
Instruction : Given a question that has multiple answers, answer the question following the instructions below:\\
\\
1. Explain step-by-step, and then provide your answer. \\
2. When providing an answer, use the format ||ANSWERS|| where ANSWERS are the answers to the given question.\\
3. Separate each answer of ANSWERS with a comma and a space.\\
\\
The format of the final answer should be given as follows:\\
\\
||ANSWER1, ANSWER2, ANSWER3||\\
\\
Now, please answer this question.\\
\\
Input : \\
\\
Question: \\
\\
\{question\}\\
\\
Answer : \\
\end{tcolorbox}
\end{footnotesize}

Moreover, to calculate the verbalize confidence, we follow the instruction format of \citet{xiong2023can}, displayed below:

\begin{footnotesize}
\begin{tcolorbox}[breakable, enhanced, top=1pt, left=1pt, right=1pt, bottom=1pt]
Instruction : Given a question that has multiple answers, answer the question and then provide the confidence in this answer, which indicates how likely you think your answer is true, following the instructions below:\\
\\
1. Keep your response as brief as possible without any explanation, and then provide your answer and confidence. \\
2. When providing an answer, use the format ||ANSWERS|| where ANSWERS are the answers to the given question.\\
3. Separate each answer of ANSWERS with a comma and a space.\\
4. The confidence should be a numerical number in the range of 0-100.\\
\\
Use the following format for the final answer and confidence:\\
\\
||ANSWER1, ANSWER2, ANSWER3||, CONFIDENCE \\
\\
Now, please answer this question. \\
\\
Question: \\
\end{tcolorbox}
\end{footnotesize}

The generated answers are then parsed and compared to the ground truth for evaluation.

\begin{table*}[htb!]
\resizebox{\textwidth}{!}{
\centering
\begin{tabular}{c|l|l|c}
\toprule
Source & Question  & LLM Answers & \# answers \\
\midrule
\textbf{AmbigQA} & Which religion has the highest population in africa? & [
                ``Christians", ``Christiany"
            ]& 2 \\
\midrule
\data & Who were the original members of the Traveling Wilburys? & [
                ``Tom Petty",
            ``Jeff Lynne",
            ``Roy Orbison",
            ``George Harrison",
            ``Bob Dylan"
            ]& 5 \\

\bottomrule
\end{tabular}
}
\caption{Examples of answers from Llama3-8b with different question types: one with ambiguity and one without.}
\label{table:ambig_example}
\end{table*}

\begin{figure*}[t]
\centering
\small
    \begin{subfigure}[b]{0.32\textwidth}
    \centering
    \small
    \includegraphics[width=1.0\linewidth]{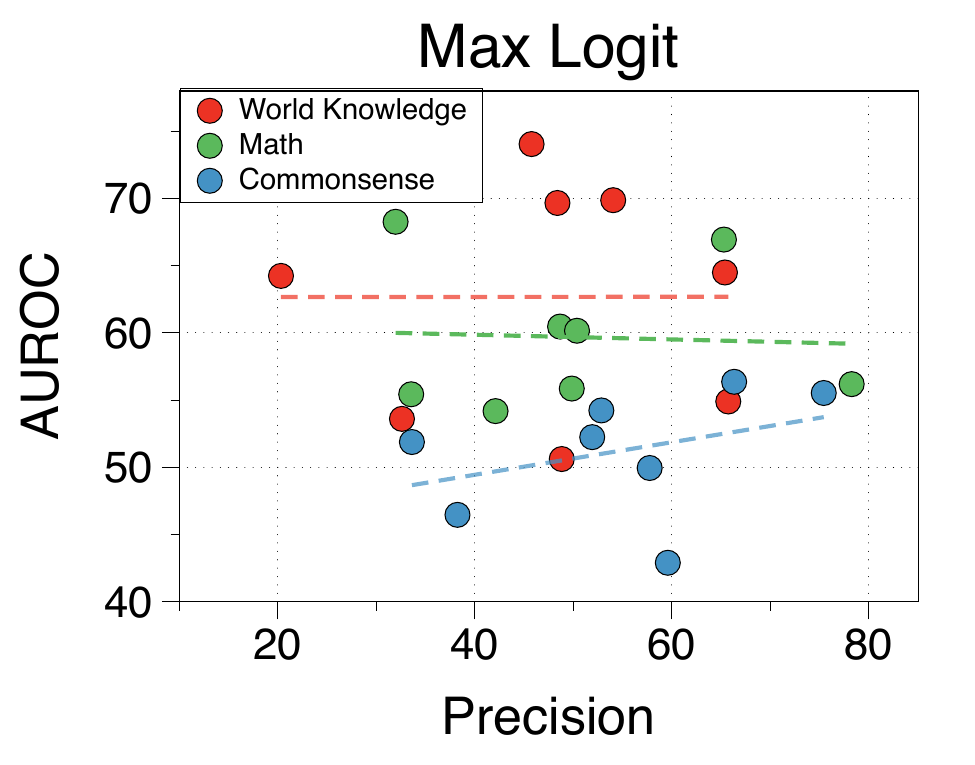}
        \caption{
        AUROC by precision for max logit
        }\label{fig:acc_logit}
    \end{subfigure}
    \hspace{-5pt}
    \begin{subfigure}[b]{0.32\textwidth}
    \centering
    \small
    \includegraphics[width=1.0\linewidth]{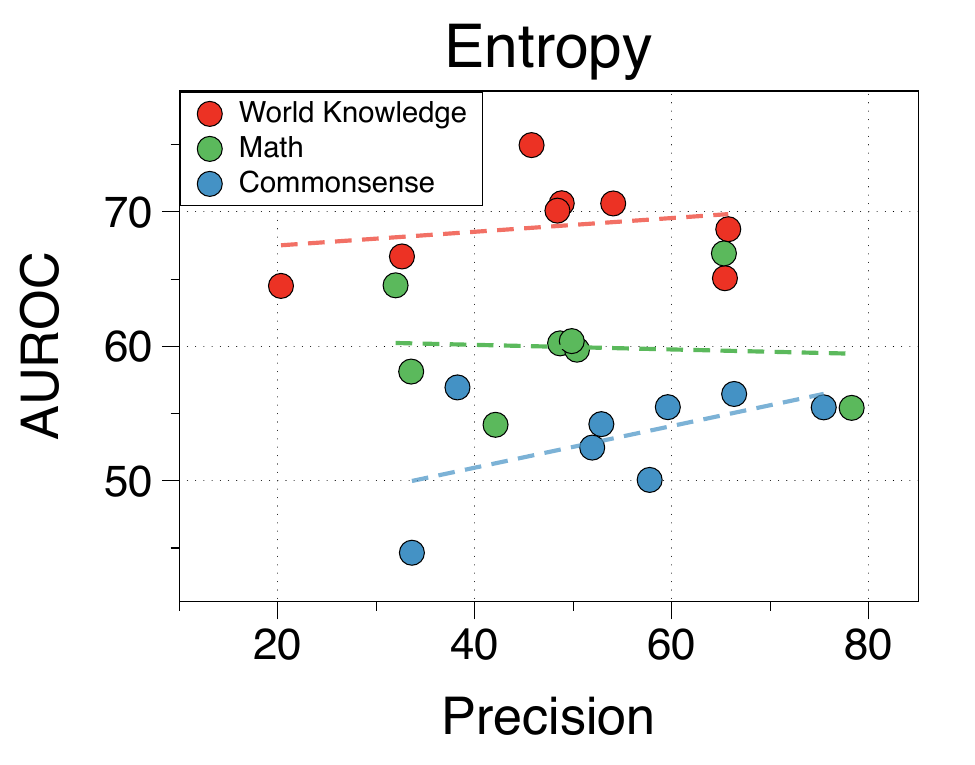}
        \caption{
         AUROC by precision for entropy
        }\label{fig:acc_entropy}
    \end{subfigure}
    \hspace{-5pt}
    \begin{subfigure}[b]{0.32\textwidth}
    \centering
    \small
    \includegraphics[width=1.0\linewidth]{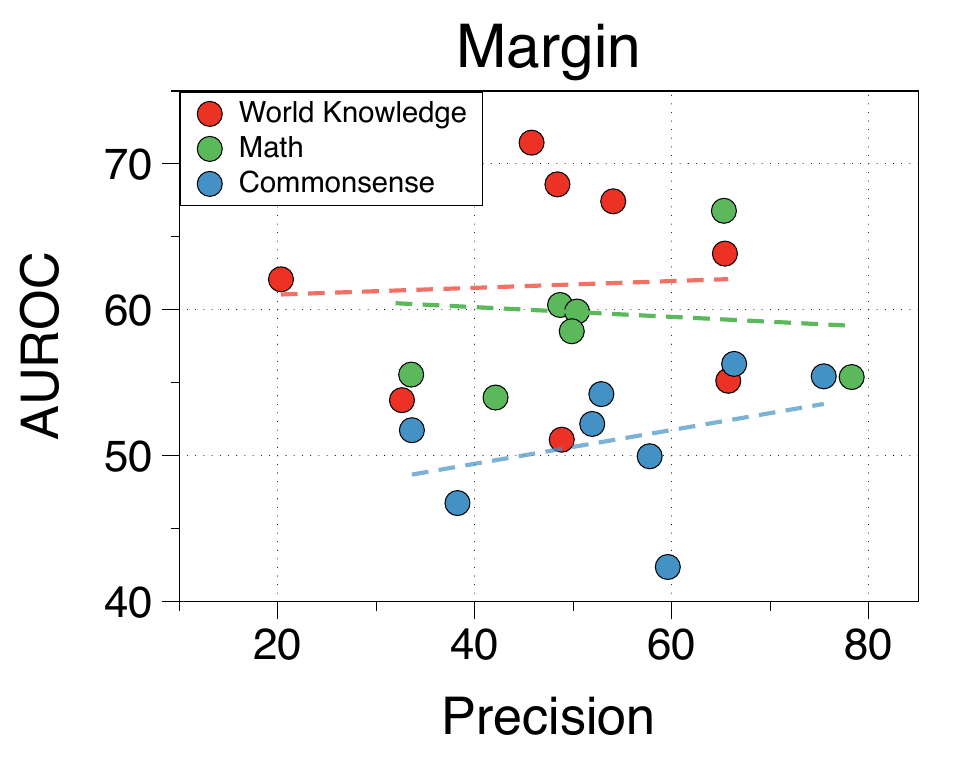}
        \caption{
         AUROC by precision for margin
        }\label{fig:acc_margin}
    \end{subfigure}
    \caption{AUROC scores by precision for \data using uncertainty quantification methods: (a) Max Softmax Logit, (b) Entropy, (c) Margin. }
    \label{fig:acc_white}
\end{figure*}
\begin{figure}[t]
\centering
\small

\includegraphics[width=1.0\linewidth]{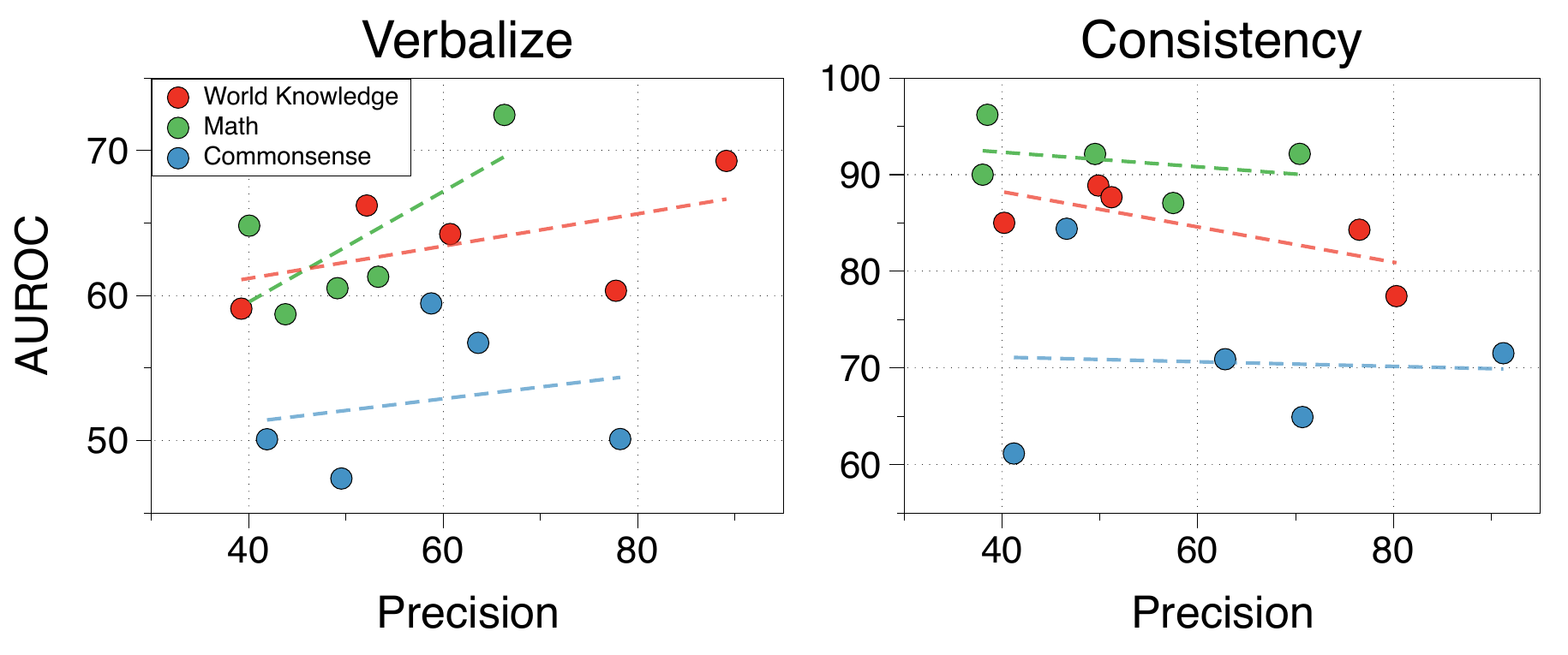}
    \caption{AUROC scores by precision for \data using uncertainty quantification methods: Verbalized Confidence and Response Consistency.}
    \label{fig:acc_black}
\end{figure}

\paragraph{Metric}

We primarily utilize \textit{accuracy} to assess correctness for single-answer and \textit{precision} for multi-answer settings.
For further analysis, we also define \textit{recall}, which is the proportion of correctly predicted answers out of all ground-truth answers. Additionally, we use the \textit{F1 score}, which is the harmonic mean of precision and recall, to provide a balanced measure of the model's accuracy in predicting multiple answers.

For the experiments with white-box LLMs, we calculate the logit-based uncertainty quantification for each answer, and the total AUROC score is calculated using all the answers. This results in the number of predictions and true labels being much higher than the number of question-answer pairs.

For the evaluation of black-box methods, since it is impossible to calculate the uncertainty for each answer, we use a threshold for the precision score. Specifically, we set the prediction label as 1 if the precision score is higher than $0.5$, as setting the threshold at $1.0$ results in too low accuracy, making the evaluation of uncertainty quantification difficult.

\section{Analysis of Model Behaviors by Question Types}
\label{appendix:qual_compare_ambig}

The multi-answer nature can arise from questions that clearly require multiple answers, as in the case of \data, or from ambiguity, even when the question requires a single answer. Here, we qualitatively analyze how the models behave differently by the existence of ambiguity in questions.  

As shown in Table~\ref{table:ambig_example}, when LLMs respond to an ambiguous question, even when notified that it is a multi-answer question, they tend to interpret it in one way and provide similar answers. In the example question, ``Which religion has the highest population in Africa?'' the multi-answer responses stem from variations in the sources that calculate population and the year of those sources, leading models to confidently provide two answers with the same meaning. On the other hand, questions from \data clearly request multiple answers, prompting the models to generate multiple distinct responses, one by one.

\section{Additional Results}

\label{appendix:more_results}

\paragraph{Internal Priority}

Table~\ref{tab:result_priority} presents the results of white-box uncertainty quantification methods using various prompt strategies. In ``priority prompting,'' we include an additional instruction that directs the model to provide responses in the order it deems most probable and common. 

As observed, using priority prompting significantly increases the quantification scores. This suggests that LLMs have internal priorities, and by encouraging them to utilize these, we can improve quantification performance. It also implies that the reduced quantification performance for \data compared to the single-answer dataset is due to the data uncertainty introduced by the multi-answer nature.

\begin{table}[t]
\centering
\resizebox{\linewidth}{!}{
\begin{tabular}{cccccccc}
\toprule[0.1em]
\multirow{2}{*}{Method} & \multicolumn{2}{c}{World Knowledge NQ} & \multicolumn{2}{c}{World Knowledge HLS} \\
\cmidrule{2-5}
& w/o priority & w priority & w/o priority & w priority  \\
\midrule
\midrule

Max Logit & 72.96 & \textbf{73.75} & 59.34 & \textbf{69.74}  \\ 
\midrule
Entropy & 73.02 & \textbf{74.89} & 60.07 & \textbf{78.61}  \\
\midrule
Margin & 70.09 & \textbf{72.46} & 57.51 & \textbf{77.61}   \\
\midrule

\bottomrule[0.1em]
\end{tabular}
}
\caption{The AUROC scores for different promptings using the Llama-3-8b model on world knowledge datasets. Priority prompting includes the additional instruction of providing answers in a way that makes the model's responses seem more probable and common.}
\label{tab:result_priority}
\end{table}

\paragraph{Recall and F1 score}

\begin{table}[t]
\centering
\resizebox{\linewidth}{!}{
\begin{tabular}{cccccccc}
\toprule[0.1em]
\multirow{2}{*}{Method} & \multicolumn{3}{c}{World Knowledge NQ} & \multicolumn{3}{c}{World Knowledge HLS} \\
\cmidrule{2-7}
& precision & recall & f1 score & precision & recall & f1 score  \\
\midrule
\midrule

Max Logit & 72.85 & 65.47 & 69.74 & 59.34 & 65.69 & 67.81   \\ 
\midrule
Entropy & 72.99 & 65.96 & 70.37 & 60.07  & 66.86 & 68.00   \\
\midrule
Margin & 70.09 & 64.64 & 73.89 & 57.51 & 65.50 & 68.19   \\
\midrule
Verbalize & 67.13 & 62.05 & 64.27 & 80.56 & 69.49 & 68.75 \\
\midrule
Consistency & \textbf{88.51} & \textbf{84.51} & \textbf{88.92} & \textbf{80.79} & \textbf{90.06} & \textbf{91.24}   \\
\midrule
\bottomrule[0.1em]
\end{tabular}
}
\caption{The AUROC scores by different correctness metrics using the Llama-3-8b model on two different world knowledge datasets.}
\label{tab:result_recall_f1}
\end{table}

To test the uncertainty quantification methods for predicting aggregated scores, we also calculate the AUROC score using recall and F1 score as true labels on the world knowledge set with the Llama3-8b model. Table~\ref{tab:result_recall_f1} shows the results of uncertainty quantification methods for predicting recall and F1 score. As we can see, regardless of the size of true labels, high AUROC scores for both recall and F1 scores are observed, with outstanding performance in response consistency. This demonstrates that uncertainty quantification methods are also related to LLMs' willingness to answer all the questions, even without explicit prompting to respond to all.

\begin{table*}[t]
\centering
\small
\resizebox{0.95\textwidth}{!}{
\begin{tabular}{cccccccccccccc}
\toprule[0.1em]
\multirow{2}{*}{Model} & \multirow{2}{*}{Method} & \multicolumn{3}{c}{World Knowledge} & \multicolumn{3}{c}{Mathematical Reasoning} & \multicolumn{3}{c}{Commonsense Reasoning} & \multicolumn{3}{c}{Overall}\\
\cmidrule{3-14}
& & single & multi & all & single & multi & all & single & multi & all & single & multi & all\\
\midrule
\midrule

\multirow{3}{*}{Qwen1.5-7b} & Max Logit & \textbf{51.60} & 42.32 & 44.28 & 60.36 & \textbf{63.25} & 62.18 & \textbf{68.64} & 55.97 & 62.21 & \textbf{60.20} & 53.85 & 56.22 \\
& Entropy & \textbf{51.16} & 42.17 & 43.86 & 60.40 & \textbf{64.12} & 62.59 & \textbf{68.77} & 55.96 & 62.30 & \textbf{60.11} & 54.08 & 56.25 \\
& Margin & \textbf{51.06} & 41.27 & 43.31 & 60.90 & \textbf{63.45} & 62.23 & \textbf{68.33} & 55.96 & 62.10 & \textbf{60.10} & 53.56 & 55.88 \\
\midrule
\multirow{3}{*}{Mistral-v02-7b} & Max Logit & 57.94 & 57.62 & \textbf{61.07} & \textbf{52.37} & 45.64 & 47.84 & \textbf{42.76} & 31.38 & 37.05 & \textbf{51.02} & 44.88 & 48.65 \\
& Entropy & 64.07 & 65.08 & \textbf{68.38} & \textbf{51.57} & 46.06 & 47.34 & \textbf{62.24} & 39.38 & 46.97 & \textbf{59.29} & 50.17 & 54.23 \\
& Margin & 58.05 & 57.81 & \textbf{61.20} & \textbf{52.15} & 46.77 & 48.29 & \textbf{43.16} & 31.42 & 37.20 & \textbf{51.12} & 45.33 & 48.90 \\
\midrule
\multirow{3}{*}{Llama3-8b} & Max Logit & \textbf{75.50} & 72.41 & 72.28 & \textbf{76.95} & 60.58 & 68.30 & \textbf{67.98} & 57.32 & 59.26 & \textbf{73.48} & 63.44 & 66.61 \\
& Entropy & \textbf{75.49} & 72.67 & 72.52 & \textbf{76.18} & 60.37 & 68.24 & \textbf{68.53} & 57.39 & 59.37 & \textbf{73.40} & 63.47 & 66.71 \\
& Margin & \textbf{74.24} & 71.43 & 71.06 & \textbf{76.32} & 60.41 & 68.31 & \textbf{67.55} & 57.34 & 59.21 & \textbf{72.70} & 63.06 & 66.20 \\
\midrule
\multirow{3}{*}{Mixtral-8x7b} & Max Logit & 69.17 & 72.83 & \textbf{76.70} & \textbf{76.08} & 53.68 & 62.06 & 45.34 & \textbf{53.29} & 46.84 & \textbf{63.53} & 59.93 & 61.87 \\
& Entropy & 83.60 & 77.89 & \textbf{84.79} & \textbf{76.27} & 60.69 & 65.06 & 60.97 & \textbf{62.53} & 61.29 & \textbf{73.61} & 67.04 & 70.38 \\
& Margin & 69.75 & 72.67 & \textbf{76.65} & \textbf{76.57} & 59.57 & 64.98 & 45.76 & \textbf{53.06} & 47.20 & \textbf{64.03} & 61.77 & 62.94 \\
\midrule
\multirow{3}{*}{Llama3-70b} & Max Logit & \textbf{76.16} & 75.98 & 75.36 & \textbf{94.29} & 80.52 & 85.78 & \textbf{78.05} & 77.95 & 77.38 & \textbf{82.83} & 78.15 & 79.51 \\
& Entropy & \textbf{76.89} & 76.09 & 75.61 & \textbf{93.84} & 79.66 & 85.11 & \textbf{78.08} & 77.40 & 77.12 & \textbf{82.93} & 77.72 & 79.28 \\
& Margin & 75.76 & \textbf{75.80} & 75.12 & \textbf{93.74} & 79.76 & 85.10 & \textbf{77.97} & 77.41 & 77.08 & \textbf{82.49} & 77.65 & 79.10 \\
\midrule
\multirow{3}{*}{Qwen1.5-72b} & Max Logit & \textbf{79.32} & 74.09 & 75.00 & \textbf{86.38} & 77.71 & 79.83 & \textbf{87.28} & 70.50 & 77.50 & \textbf{84.32} & 74.10 & 77.44 \\
& Entropy & \textbf{79.56} & 74.44 & 75.28 & \textbf{85.56} & 77.47 & 79.38 & \textbf{87.30} & 70.58 & 77.55 & \textbf{84.14} & 74.17 & 77.40 \\
& Margin & \textbf{78.63} & 73.77 & 74.58 & \textbf{85.63} & 77.28 & 79.36 & \textbf{87.32} & 70.59 & 77.54 & \textbf{83.86} & 73.88 & 77.16 \\
\midrule
\multirow{3}{*}{Average} & Max Logit & \textbf{68.28} & 65.87 & 67.45 & \textbf{74.41} & 63.56 & 67.66 & \textbf{65.01} & 57.73 & 60.04 & \textbf{69.23} & 62.39 & 65.05 \\
& Entropy & \textbf{71.79} & 68.06 & 70.07 & \textbf{73.97} & 64.73 & 67.95 & \textbf{70.98} & 60.54 & 64.10 & \textbf{72.25} & 64.44 & 67.38 \\
& Margin & \textbf{67.91} & 65.46 & 66.99 & \textbf{74.22} & 64.54 & 68.04 & \textbf{65.02} & 57.63 & 60.06 & \textbf{69.05} & 62.54 & 65.03 \\
\midrule
\bottomrule[0.1em]
\end{tabular}
}
\caption{The AUPRC scores obtained using white-box based uncertainty quantification models on different tasks, methods, and models. A high score indicates high quantification performance. ``single'' refers to the results on a QA set with only single answers, ``multi'' refers to the \data, and ``all'' refers to the combination of these two datasets, which includes both single and multi-answer sets.}
\label{tab:result_white_auprc}
\end{table*}
\begin{table*}[t]
\centering
\small
\resizebox{0.95\textwidth}{!}{
\begin{tabular}{cccccccccccccc}
\toprule
\multirow{2}{*}{Model} & \multirow{2}{*}{Method} & \multicolumn{3}{c}{World Knowledge} & \multicolumn{3}{c}{Mathematical Reasoning} & \multicolumn{3}{c}{Commonsense Reasoning} & \multicolumn{3}{c}{Overall} \\
\cmidrule{3-14}
& & single & multi & all & single & multi & all & single & multi & all & single & multi & all \\
\midrule
\midrule

\multirow{2}{*}{Qwen1.5-7b} & Verbalize & \textbf{48.41} & 39.70 & 34.82 & 32.49 & \textbf{36.15} & 35.53 & 58.46 & \textbf{59.02} & 58.31 & \textbf{46.45} & 44.96 & 42.89 \\
 & Consistency & \textbf{41.64} & 36.77 & 39.15 & \textbf{95.01} & 90.51 & 86.61 & \textbf{66.23} & 58.64 & 61.40 & \textbf{67.63} & 61.97 & 62.39 \\
\midrule
\multirow{2}{*}{Mistral-v02-7b} & Verbalize & 55.37 & \textbf{63.44} & 55.01 & \textbf{54.26} & 44.18 & 48.72 & \textbf{62.70} & 39.43 & 49.82 & \textbf{57.44} & 49.02 & 51.18 \\
 & Consistency & 57.64 & \textbf{78.78} & 59.82 & \textbf{89.19} & 84.01 & 79.59 & \textbf{61.02} & 58.28 & 60.44 & 69.28 & \textbf{73.69} & 66.62 \\
\midrule
\multirow{2}{*}{Llama3-8b} & Verbalize & 63.07 & \textbf{68.94} & 58.37 & \textbf{80.98} & 62.28 & 75.60 & \textbf{65.98} & 41.73 & 61.82 & \textbf{70.01} & 57.65 & 65.26 \\
 & Consistency & 72.70 & 73.46 & \textbf{75.02} & \textbf{97.84} & 89.43 & 94.27 & \textbf{62.47} & 61.25 & 61.82 & \textbf{77.67} & 74.71 & 77.04 \\
\midrule
\multirow{2}{*}{GPT-3.5} & Verbalize & 60.62 & \textbf{67.02} & 64.45 & \textbf{75.14} & 53.70 & 67.62 & \textbf{72.00} & 53.63 & 66.89 & \textbf{69.25} & 58.12 & 66.32 \\
 & Consistency & 78.23 & \textbf{79.32} & 78.71 & \textbf{95.09} & 84.78 & 91.27 & \textbf{77.49} & 76.11 & 75.52 & \textbf{83.60} & 80.07 & 81.83 \\
\midrule
\multirow{2}{*}{GPT-4} & Verbalize & 78.82 & \textbf{85.73} & 76.24 & \textbf{95.16} & 81.07 & 92.97 & 75.82 & 83.59 & \textbf{85.24} & 83.27 & 83.46 & \textbf{84.82} \\
 & Consistency & 70.26 & \textbf{78.03} & 75.45 & \textbf{97.07} & 88.78 & 96.12 & 84.90 & 83.72 & \textbf{85.04} & 84.08 & 83.51 & \textbf{85.54} \\
\midrule
\multirow{2}{*}{Average} & Verbalize & 61.26 & \textbf{64.97} & 57.78 & \textbf{67.61} & 55.48 & 64.09 & \textbf{66.99} & 55.48 & 64.42 & \textbf{65.29} & 58.64 & 62.09 \\
 & Consistency & 64.09 & \textbf{69.27} & 65.63 & \textbf{94.84} & 87.50 & 89.57 & \textbf{70.42} & 67.60 & 68.84 & \textbf{76.45} & 74.79 & 74.68 \\
\midrule
\bottomrule
\end{tabular}
}
\caption{The AUPRC scores obtained using black-box based uncertainty quantification models on different tasks, methods, and models. A high score indicates high quantification performance. ``single'' refers to the results on a QA set with only single answers, ``multi'' refers to the \data, and ``all'' refers to the combination of these two datasets, which includes both single and multi-answer sets. }
\label{tab:result_black_auprc}
\end{table*}

\paragraph{Model Capability and Uncertainty Quantification}

Figure~\ref{fig:acc_white} shows the correlation between the precision score, defined as the correctness of the LLMs, and the uncertainty quantification using white-box-based methods. The results demonstrate that there is little correlation with the world knowledge dataset, a higher correlation with the commonsense dataset, and no significant correlation with the mathematical reasoning dataset. This indicates that higher capability does not always correlate with the performance of uncertainty quantification, as it is more dependent on the model and task. 

This trend differs for black-box-based methods. As shown in Figure~\ref{fig:acc_black}, there is a high correlation between the precision and AUROC for verbalized confidence. This implies that as LLMs become more capable and possess better knowledge, they also become more capable of accurately predicting their confidence. Additionally, response consistency shows the opposite trend of verbalized confidence, as the performance of using response consistency shows a slightly negative correlation with the precision score. This may be due to the fact that as accuracy increases, response consistency becomes less effective at distinguishing between correct and incorrect answers, reducing its utility as a confidence measure.





\paragraph{Different Metric}
Table~\ref{tab:result_white_auprc} presents the AUPRC scores for the uncertainty quantification methods for white-box LLMs, with the positive label set as 1. For the AUPRC score, the scores mostly decrease for multi-answer sets in all tasks, showing the impact of data uncertainty induced by the multi-answer nature. Still, entropy serves as the best uncertainty quantification method among the white-box methods. Additionally, these trends differ significantly by model, but the AUPRC scores generally increase as the model size grows, as evidenced by the performance on Mixtral, Llama3-70b, and Qwen1.5-72b.

Table~\ref{tab:result_black_auprc} demonstrates the AUPRC scores for black-box methods. Unlike white-box-based methods, the scores mostly decrease for multi-answer sets in reasoning tasks, while the scores somewhat increase for world knowledge tasks. This implies that black-box-based methods are less affected by the presence of data uncertainty. Nonetheless, consistency serves as a strong baseline, especially for the multi-answer datasets, showing the efficacy of estimating model uncertainty under data uncertainty.

\begin{figure*}[htb!]
\centering
\small

\includegraphics[width=1.0\linewidth]{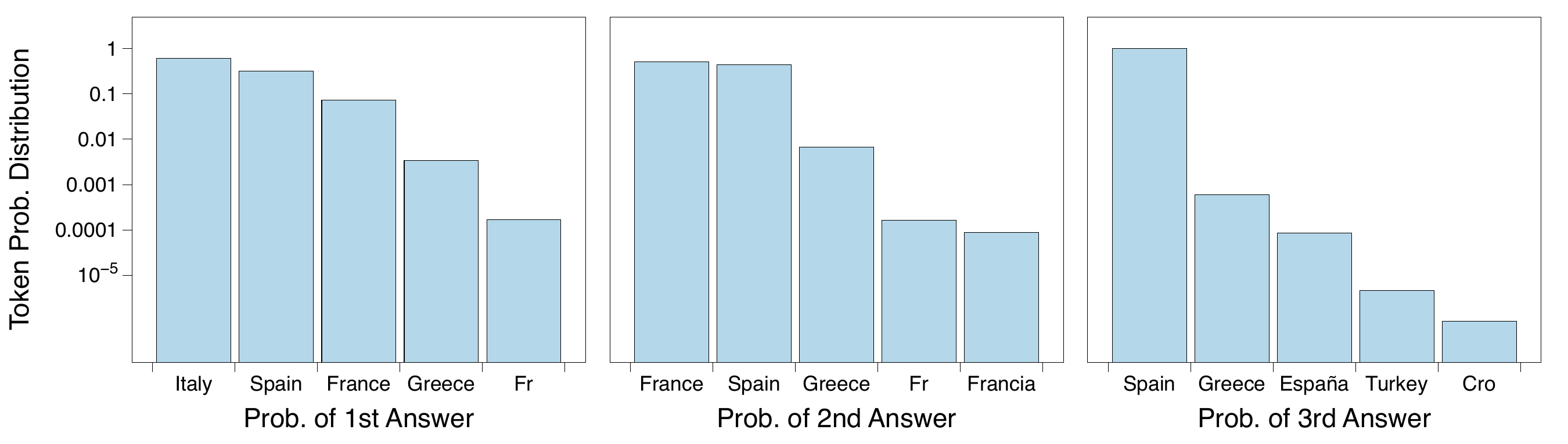}
    \caption{Examples of the probability distribution for each answer in the response from the Mistral-v02 model to the question, ``Which countries have coastlines on the Mediterranean Sea?"}
    \label{fig:qual_white}
\end{figure*}


\begin{table*}[htb!]
\resizebox{\textwidth}{!}{
\centering
\begin{tabular}{l|p{15cm}|c|c}
\toprule
Question  & Answers & C & P \\
\midrule
\multirow{5}{8cm}{List members of the Ivy League, a specific group of prestigious universities predominantly in the Northeastern United States.} & [
                ``Brown",
            ``Columbia",
                ``Cornell",
                ``Dartmouth",
                ``"Harvard",
                ``Penn",
                ``Princeton",
                ``Yale"
            ]&  \multirow{5}{*}{0.83} &  \multirow{5}{*}{96.42} \\
            & [
                ``Brown",
                ``Columbia",
                ``Cornerstone",
                ``Dartmouth",
                ``Harvard",
                ``Penn",
                ``Princeton",
                ``Yale"
            ]& & \\
            &[
                ``Brown",
                ``Columbia",
                ``Cornerstone",
                ``Dartmouth",
                ``Harvard",
                ``Penn",
                ``Princeton",
                ``Yale"
            ]&& \\
            &[
                ``Brown",
                ``Columbia",
                ``Cornerstone",
                ``Dartmouth",
                ``Harvard",
                ``Penn",
                ``Princeton",
                ``Yale"
            ]&& \\
            &[
                ``Brown",
                ``Columbia",
                ``Cornerstone",
                ``Dartmouth",
                ``Harvard",
                ``Penn",
                ``Princeton",
                ``Yale"
            ]& & \\ 

\midrule

\multirow{5}{8cm}{What are the names of the actresses who voiced the muses in Disney's ``Hercules"?} & [
                ``Jennifer Lewis",
                ``Liz Callaway",
                ``Susan Egan"
            ] &  \multirow{5}{*}{0.19} &  \multirow{5}{*}{4.00} \\
             &[
                ``Jennifer Love Hewitt",
                ``Rene Russo",
                ``Patrick Dempsey",
                ``Teri Hatcher",
                ``Jennifer Coolidge"
            ]& & \\ 
             &[
                ``Meg Ryan",
                ``Christy Carlson Romano",
                ``Roz Ryan",
                ``Lynn Hollander",
                ``Joseph Ann Sullivan"
            ]& & \\ 
             &[
                ``Meg Ryan",
                ``Lindsay Lohan",
                ``Rene Russo",
                ``Patrick Stewart",
                ``Jennifer Love Hewitt",
                ``Christy Carlson Romano",
                ``Zoe Shanahan"
            ]& & \\ 
             &[
                ``Meg Ryan",
                ``Liz Callaway",
                ``Jennifer Lewis"
            ]& & \\

\midrule

\multirow{5}{8cm}{Shawna's father is x times as old as Shawna. Shawna is currently three times as old as Aliya. If Aliya is 3 years old, determine all possible values of x that would make Shawna's father's age more than 30 but less than 60.} & [
                4,
                5
            ] &  \multirow{5}{*}{0.79} &  \multirow{5}{*}{88.00} \\
 &[
                4,
                5,
                6,
                7
            ]& & \\ 
 &[
                3,
                4,
                5
            ]& & \\ 
 &[
                4,
                5
            ]& & \\ 
 &[
                4,
                5,
                6
            ]& & \\

\bottomrule
\end{tabular}
}
\caption{Examples of sampling multiple answers for each question using Llama3-8b, along with its precision score (P) and uncertainty estimation using consistency (C).}
\label{table:consistency_example}
\end{table*}

\section{Qualitative Results}
\label{appendix:qualitative}

Figure~\ref{fig:qual_white} shows examples of token probability distributions when LLMs are asked a multi-answer question. As the question has multiple answers, including ``Italy,'' ``Spain,'' and ``Greece,'' we can clearly see that the token probability is distributed among the answers, which could be interpreted as data uncertainty. However, the probability of the fifth token remains below 0.001, implying an internal priority within the LLMs.

Table~\ref{table:consistency_example} presents examples of using response consistency for uncertainty quantification. As observed, when LLMs demonstrate higher consistency scores, they typically show high correctness, which corresponds to high precision in this case. However, this could be attributed to the fact that the answers are in a short format, allowing for fine-grained evaluation of consistency.

\section{More Related Work}
\label{appendix:more_related_work}


\subsection{Question and Answering Datasets}

To test the diverse abilities of language models, multiple open-domain question answering (ODQA) datasets~\citep{kwiatkowski-etal-2019-natural, min2020ambigqa, joshi-etal-2017-triviaqa, boratko2020protoqa, zhu-etal-2020-question, lin2021truthfulqa} have been proposed. These datasets involve the task of answering any factual question. Early benchmarks created open-ended questions based on evidence from certain Wikipedia paragraphs~\citep{chen-etal-2017-reading}, framing the task as reading comprehension. Due to the development of LLMs, these QA tasks are sometimes tested using only the LLMs without evidence~\citep{jiang2023mistral}, assuming a deterministic single answer for each question.

Among these datasets, some datasets contain multi-answer question-answer pairs~\citep{joshi-etal-2017-triviaqa, kwiatkowski-etal-2019-natural, min2020ambigqa}. Specifically, 
TriviaQA~\citep{joshi-etal-2017-triviaqa} includes some questions with multiple answers that can be found in documents. Natural Questions~(NQ)~\citep{kwiatkowski-etal-2019-natural} collects numerous user queries from Google, some of which require multiple answers. AmbigQA~\citep{min2020ambigqa} also contains multi-answer question tasks arising from ambiguity,  with most of the questions based on the NQ dataset. Additionally, there is a line of research focusing on multi-span reading comprehension tasks~\citep{li-etal-2022-multispanqa, zhu-etal-2020-question, malaviya2023quest}.

\subsection{Uncertainty Benchmarks}

There is also a line of research focused on developing uncertainty benchmarks specifically designed to evaluate the uncertainty quantification of large foundation models~\citep{ye2024benchmarking,wang2024ubench, chandu2024certainly}. These datasets primarily either incorporate conformal prediction methods to create uncertainty benchmarks~\citep{ye2024benchmarking}, or they assess uncertainty through model responses~\citep{wang2024ubench}. Additionally, a benchmark has been proposed that accounts for two sources of uncertainty in the vision-language domain~\citep{chandu2024certainly}.

\subsection{Uncertainty Decomposition for LLMs}

Uncertainty decomposition has been widely studied in machine learning, particularly in computer vision domains~\citep{hullermeier2021aleatoric, valdenegro2022deeper, he2023survey}.
Recently, efforts have been made to decompose model uncertainty and data uncertainty in the context of LLMs~\citep{hou2023decomposing, cole2023selectively}. These methods typically involve a clarification stage, a process that adds more details to the question to eliminate data uncertainty caused by the ambiguity of the user's question. 
However, even though the questions are unambiguous, there are many cases where data uncertainty still exists, and users require multiple answers, motivating us to investigate uncertainty quantification under the existence of data uncertainty without clarification.













\end{document}